\documentclass[10pt,twocolumn,letterpaper]{article}

\usepackage[margin=0.75in]{geometry}
\usepackage{times}
\usepackage{helvet}
\usepackage{courier}
\usepackage{url}
\usepackage{graphicx}
\usepackage{amsmath,amssymb}
\usepackage{booktabs}
\usepackage{multirow}
\usepackage{tabularx}
\usepackage{microtype}
\usepackage{tikz}
\usetikzlibrary{arrows.meta,positioning,fit,calc}
\usepackage[numbers,sort&compress]{natbib}

\newcommand{\placeholderimage}[2]{%
\IfFileExists{#1}{\includegraphics[width=#2]{#1}}{%
\fbox{\parbox[c][#2][c]{#2}{\centering\scriptsize Missing image}}}%
}
\newcommand{\imgcell}[3]{%
\begin{minipage}[t]{#1}\centering
\placeholderimage{#2}{\linewidth}\\[-0.2em]
{\scriptsize #3}
\end{minipage}%
}

\newcommand{\imgcellb}[3]{%
\begin{minipage}[b]{#1}\centering
\placeholderimage{#2}{\linewidth}\\[-0.2em]
{\scriptsize #3}
\end{minipage}%
}

\title{Compressing Image Style Training into a Single Model Forward}
\author{Zhongjie Duan \quad Yingda Chen\\
ModelScope Team, Alibaba Group\\
\texttt{duanzhongjie.dzj@alibaba-inc.com}}
\date{}

\begin{document}
\maketitle

\begin{abstract}
Diffusion-based style transfer must balance inference efficiency with stylization fidelity. Adapter-based methods are efficient, but they inject style as an external condition and can either weaken reference-specific appearance or copy reference semantics into the generated image. Optimization-based personalization methods such as LoRA internalize style more effectively, but require a separate training process for every new style. We introduce i2L (image-to-LoRA), a framework that amortizes style LoRA training into a single forward pass. Given one or more reference images, i2L predicts LoRA weights for a text-to-image model, enabling immediate style instantiation without per-style optimization. The architecture combines an image encoder, learnable LoRA queries, and compressed decoding heads that generate adapted matrices. Training on semantically diverse style pairs encourages the predictor to preserve appearance cues while suppressing reference-content copying. Experiments on Z-Image, FLUX.2, and Hidream-O1 show that i2L improves style fidelity, prompt alignment, and perceptual quality over existing baselines. Because i2L produces explicit LoRA weights, it also supports asymmetric classifier-free guidance, multi-reference style fusion, and composition with controllable-generation modules.
\end{abstract}

\section{Introduction}

Image style transfer aims to synthesize images whose content follows a user specification while their visual appearance follows a reference style. Early neural style transfer methods optimized convolutional feature statistics, and subsequent arbitrary style transfer models learned feed-forward stylizers for efficient inference \cite{gatys2016image,huang2017adain}. However, these methods typically represent style through local texture, color palette, and brushstroke statistics. They are less effective when style involves high-level composition, object deformation, material priors, lighting, typography, or the distinctive visual language of a creator or collection.

Large-scale diffusion models \cite{rombach2022ldm,zimage2025,hidream2025} have substantially expanded the design space for style transfer. Text-to-image diffusion models provide strong natural-image priors, flexible text control, and rich internal representations that jointly encode semantics, layout, and appearance \cite{ho2020ddpm,rombach2022ldm}. Existing diffusion-based style transfer methods can be broadly grouped into two categories. The first learns an external conditioning module, such as ControlNet \cite{zhang2023controlnet}, T2I-Adapter \cite{mou2023t2iadapter}, or IP-Adapter \cite{ye2023ipadapter}, to inject reference-image features or auxiliary controls into a diffusion model. These adapter-based systems are appealing because, after training, they require only a single inference pass and support arbitrary references. However, because the style signal remains an auxiliary condition rather than an internal component of the generator, such methods often suffer from weak style fidelity, prompt-reference conflict, and semantic leakage from the reference image.

The second category internalizes the reference style by optimizing embeddings or model parameters for a specific concept or style. Textual Inversion \cite{gal2023textual} learns new token embeddings, DreamBooth \cite{ruiz2023dreambooth} fine-tunes the generator to associate a rare token with a subject, and LoRA \cite{hu2022lora} adapts selected layers with low-rank residual matrices. LoRA is particularly appealing for visual style transfer because it offers a favorable trade-off between parameter efficiency and expressivity: a style can be represented as a compact set of low-rank weight updates and reused with the base diffusion model at inference time. This expressivity, however, requires per-style optimization over many diffusion training steps, making LoRA-based stylization slow, expensive, and poorly suited to interactive or large-scale deployment.

\begin{figure}[t]
    \centering
    \includegraphics[width=0.98\linewidth]{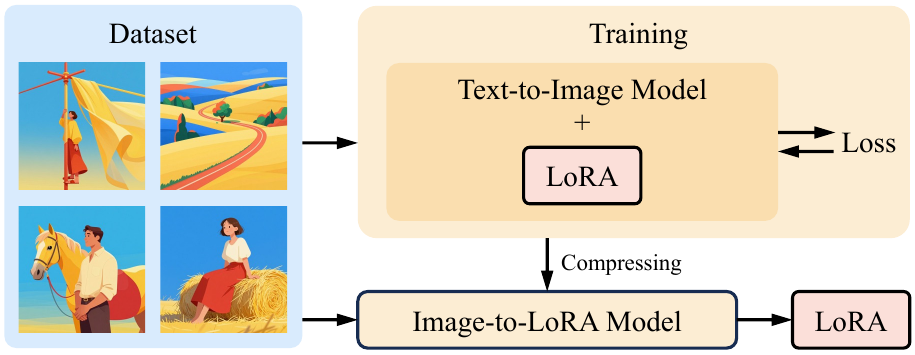}
    \caption{The workflow of Image-to-LoRA.}
    \label{fig:workflow}
\end{figure}

This work asks whether the optimization process used to train a style LoRA can be collapsed into a single model forward pass. As illustrated in Figure \ref{fig:workflow}, we propose \emph{i2L}, an image-to-LoRA architecture that maps one or more reference images directly to LoRA weights. Rather than using the reference only as an auxiliary condition, i2L predicts the same form of internal model update produced by conventional LoRA training. The expensive per-style training loop is replaced by a meta-model trained once over many style-content pairs; at test time, a new style is instantiated by forwarding its references through i2L.

The i2L architecture consists of an image encoder, a transformer with learnable LoRA queries, and compressed linear heads that decode query states into LoRA rows or columns. By aligning queries with the row-and-column structure of LoRA matrices, the predictor scales to multiple ranks and adapted layers while remaining compact. We train i2L end-to-end through frozen text-to-image backbones with the standard flow-matching objective \cite{lipman2023flowmatching,esser2024scaling}, updating only the image-to-LoRA network. To reduce reference-content copying, we train on MegaStyle-1M \cite{gao2026megastyle}, whose style-related image pairs are semantically diverse and thus encourage style preservation rather than semantic leakage. We instantiate i2L on Z-Image \cite{zimage2025}, FLUX.2 \cite{flux-2-2025}, and Hidream-O1 \cite{hidream2025}, where it improves style fidelity and prompt alignment over baseline methods. Because i2L predicts explicit LoRA weights, it further enables asymmetric classifier-free guidance \cite{ho2022classifierfree}: the positive branch uses the reference-image LoRA, whereas the negative branch uses a gray-image LoRA, strengthening stylization without additional training. The i2L models for the three base backbones are released publicly \footnote{\url{https://modelscope.cn/models/DiffSynth-Studio/ZImage-i2L-v2}}\footnote{\url{https://modelscope.cn/models/DiffSynth-Studio/KleinBase4B-i2L-v2}}\footnote{\url{https://modelscope.cn/models/DiffSynth-Studio/HidreamO1-i2L-v2}}, and the source code will be released in DiffSynth-Studio \footnote{\url{https://github.com/modelscope/DiffSynth-Studio}}.

Our contributions are summarized as follows:
\begin{itemize}
    \item We formulate style transfer as direct prediction of generator weight updates, introducing i2L to amortize per-style LoRA optimization into a single forward pass from reference images.
    \item We design a LoRA-structured predictor that uses learnable row-and-column queries with compressed decoding heads, enabling scalable generation of many layer-specific LoRA matrices.
    \item We demonstrate that explicit predicted LoRAs improve style fidelity across Z-Image, FLUX.2, and Hidream-O1, while naturally supporting asymmetric guidance, multi-reference style fusion, and composition with controllable-generation modules.
\end{itemize}

\section{Related Work}

\paragraph{Neural and diffusion-based style transfer.}
Classical neural style transfer formulates stylization as matching content features and style statistics in a pretrained network \cite{gatys2016image}. Feed-forward arbitrary style transfer methods, including AdaIN \cite{huang2017adain} and transformer-based stylizers \cite{deng2022stytr2}, improve efficiency and generalization to unseen styles. However, their reliance on discriminative features often limits their ability to capture semantic or compositional aspects of style. Diffusion models \cite{ho2020ddpm,rombach2022ldm} provide stronger generative priors and have become a common foundation for recent stylization systems. Training-free methods manipulate inversion trajectories, attention maps, or hidden features, as in StyleID and $Z^*$ \cite{chung2024styleid,deng2024zstar}; trained systems instead learn style-aware conditions or adapters \cite{gao2024styleshot}. i2L follows a different route: it neither adjusts sampling internals for each input nor represents style only as an external condition. Instead, it predicts generator weight updates that encode style within the diffusion model.

\paragraph{Adapter-based reference conditioning.}
Adapter-based methods learn modules that connect image encoders to a frozen text-to-image model. IP-Adapter-style designs are efficient because they decouple image-prompt features from text features and support arbitrary reference images at test time without per-style optimization \cite{ye2023ipadapter}. Related ideas appear in ControlNet \cite{zhang2023controlnet}, StyleCrafter \cite{liu2023stylecrafter}, and other reference-conditioned diffusion pipelines. A central limitation is that the frozen generator must receive the entire reference style through conditioning tokens or feature injections. When the target prompt differs substantially from the reference image, this bottleneck often leads to partial stylization or unwanted copying of reference semantics. i2L retains the test-time convenience of adapters but changes the output space: instead of predicting conditioning features, it predicts a LoRA model that directly modulates generator weights.

\begin{figure*}[htbp]
    \centering
    \includegraphics[width=0.98\textwidth]{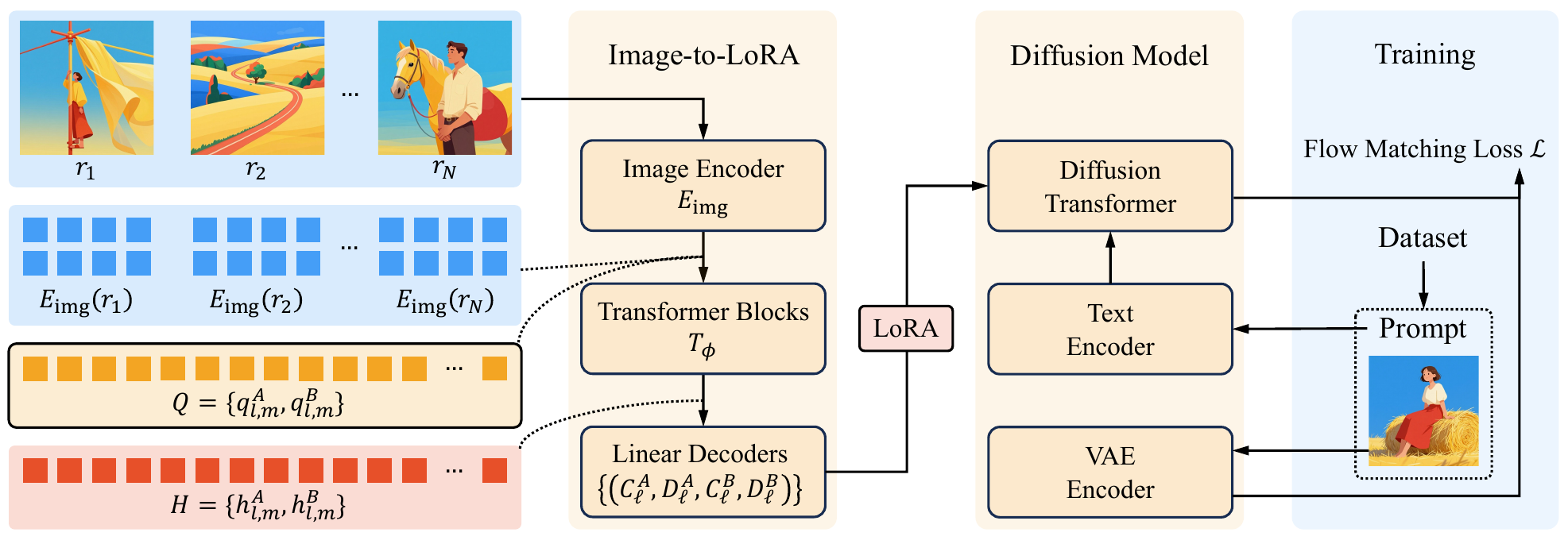}
    \caption{Overview of the proposed i2L training pipeline. Reference images are encoded by an image encoder and fused with structured LoRA queries through a stack of transformer blocks. Compressed linear heads decode LoRA query states into LoRA matrices, which are inserted into the text-to-image model. The standard flow-matching loss is back-propagated through the LoRA application to train the i2L model end-to-end.}
    \label{fig:i2l_architecture}
\end{figure*}

\paragraph{Personalization and lightweight fine-tuning.}
Personalized generation methods learn compact representations of a subject, identity, or style from a small set of examples. Textual Inversion \cite{gal2023textual} optimizes a new token embedding that activates the desired concept, while DreamBooth \cite{ruiz2023dreambooth} fine-tunes model weights to bind a rare token to a subject while preserving the model prior. LoRA \cite{hu2022lora} introduces trainable low-rank matrices into existing layers, enabling parameter-efficient adaptation of large models. For style transfer, LoRA is especially attractive because weight-space updates can capture global visual regularities more faithfully than a single token embedding. However, conventional LoRA requires iterative optimization for each new style. Our work can be viewed as amortized LoRA personalization: after meta-training, the model predicts a style-specific LoRA from images in one forward pass.

\paragraph{Hypernetworks and weight generation.}
Generating neural network weights with another network has a long history in hypernetworks and dynamic parameter prediction \cite{chauhan2024brief,jia2016dynamic}. i2L follows this paradigm, but diffusion LoRA generation introduces a pronounced scale mismatch: modern diffusion transformers contain many adapted projections, whereas the reference signal may consist of only a few images. We therefore avoid generating all adapted weights from a single pooled embedding. Instead, structured LoRA queries correspond to individual rows or columns of LoRA matrices, and per-layer compressed linear heads decode the final weights. This design keeps weight generation scalable without forcing all layers to share a generic output head.

\paragraph{Style datasets and semantic leakage.}
Reference-based stylization requires separating style from content, yet many image-text corpora entangle the two. When reference and target images depict similar content, a model can reduce its training loss by copying semantic attributes rather than learning style. This leads to semantic leakage: a cat reference may make generated dogs appear cat-like, or a portrait reference may impose identity onto unrelated prompts. MegaStyle-1M mitigates this issue by constructing large-scale stylistic correspondences with diverse content \cite{gao2026megastyle}. We adopt the same principle and train i2L with style-consistent but content-disjoint examples, encouraging the predicted LoRA to preserve appearance while respecting target semantics.

\section{Methodology}

Figure~\ref{fig:i2l_architecture} illustrates the i2L pipeline. Given a set of reference images $\mathcal{R}=\{r_i\}_{i=1}^{N}$ sharing a visual style, we predict a LoRA parameter set $\Delta\Theta_{\mathcal{R}}$ for a text-to-image diffusion model $\epsilon_{\theta}$. The predicted LoRA encodes the style of $\mathcal{R}$ while preserving content control through the text prompt. Unlike adapter methods such as ControlNet \cite{zhang2023controlnet} and IP-Adapter \cite{ye2023ipadapter}, which inject external controls or image features throughout generation, i2L predicts LoRA weights once and then follows the standard generation pipeline with the adapted model.

\subsection{Image-to-LoRA Architecture}

\paragraph{Image encoding.}
Each reference image is processed by a SigLIP2 image encoder $E_{\mathrm{img}}$ \cite{tschannen2025siglip2}. We retain patch-level embeddings rather than a single pooled token, as style may be distributed across local texture, palette, composition, and object-independent visual motifs. For $N$ reference images, the resulting image tokens are concatenated as
\begin{equation}
    Z_{\mathrm{img}} = \mathrm{Concat}\left(E_{\mathrm{img}}(r_1), \ldots, E_{\mathrm{img}}(r_N)\right).
\end{equation}
The encoder remains frozen throughout training, stabilizing optimization and preventing the image representation from drifting toward generator-specific shortcuts.

\paragraph{LoRA queries.}
Assume LoRA is inserted into $L$ selected linear layers $\{W_{\ell}\}_{\ell=1}^L$ of the diffusion backbone. For layer $\ell$, a standard rank-$k$ LoRA update is
\begin{equation}
    W_{\ell}' = W_{\ell} + \alpha_{\ell} B_{\ell} A_{\ell},
\end{equation}
where $A_{\ell}\in\mathbb{R}^{k\times d_{\ell}^{\mathrm{in}}}$, $B_{\ell}\in\mathbb{R}^{d_{\ell}^{\mathrm{out}}\times k}$, and $\alpha_{\ell}$ is a scaling factor. i2L parameterizes these matrices with learnable query embeddings. Each query corresponds to one row of $A_{\ell}$ or one column of $B_{\ell}$. The total number of LoRA queries is therefore $2kL$: for every adapted layer, $k$ queries generate the rows of $A_{\ell}$ and $k$ queries generate the columns of $B_{\ell}$. We denote the query set by $Q=\{q_{\ell,m}^{A}, q_{\ell,m}^{B}\}$.

\paragraph{Transformer aggregation.}
The image tokens and LoRA queries are concatenated and passed through a transformer \cite{vaswani2017attention} model $T_{\phi}$ composed of single-stream transformer blocks:
\begin{equation}
    H = T_{\phi}\left([Q; Z_{\mathrm{img}}]\right).
\end{equation}
Only the output states corresponding to LoRA queries are decoded into weights. Through self-attention, each query can attend to all reference-image tokens and to other LoRA queries. This enables the predictor to coordinate updates across layers and ranks, which is important because style is distributed across multiple projections in the diffusion backbone.

\paragraph{Compressed linear decoding.}
Directly mapping each query state to a full LoRA row or column with an independent large linear layer would make the predictor prohibitively large. We therefore use compressed linear decoders for each LoRA matrix type and layer. Given a query hidden state $h_{\ell,m}^{A}$, the corresponding row of $A_{\ell}$ is generated as
\begin{equation}
    A_{\ell}[m,:] = D_{\ell}^{A} C_{\ell}^{A} h_{\ell,m}^{A},
\end{equation}
where $C_{\ell}^{A}$ reduces dimensionality and $D_{\ell}^{A}$ expands to $d_{\ell}^{\mathrm{in}}$. Similarly, $h_{\ell,m}^{B}$ is decoded into the $m$-th column of $B_{\ell}$ with a separate compressed linear head. The predictor uses $2L$ compressed decoders in total, one for $A_{\ell}$ and one for $B_{\ell}$ at each adapted layer. This factorized decoding keeps the parameter count manageable while preserving layer-specific output dimensions.

\subsection{Training Objective}

We train i2L by differentiating through the frozen diffusion backbone under the standard flow-matching formulation for generative modeling \cite{lipman2023flowmatching}. For a target image $x_1$, text prompt $c$, and Gaussian noise $x_0\sim\mathcal{N}(0,I)$, flow matching constructs the interpolated latent
\begin{equation}
    x_t = (1-t)x_0 + t x_1,
\end{equation}
with target velocity $u_t=x_1-x_0$. Given reference images $\mathcal{R}$, i2L predicts LoRA weights $\Delta\Theta_{\mathcal{R}}=G_{\phi}(\mathcal{R})$ and inserts them into the frozen backbone. The training loss is
\begin{equation}
    \mathcal{L}_{\mathrm{FM}} = \mathbb{E}_{x_0,x_1,t,c,\mathcal{R}}\left[\left\|v_{\theta+\Delta\Theta_{\mathcal{R}}}(x_t,t,c)-u_t\right\|_2^2\right].
\end{equation}
Gradients pass through the LoRA application and update only the i2L parameters $\phi$; both the SigLIP2 encoder and the base text-to-image model remain frozen. The predictor therefore learns weight updates that allow the frozen generator to model target images in the desired style under standard diffusion supervision.

\subsection{Style-Disentangled Data Construction}

Training on ordinary image-text pairs can encourage the predicted LoRA to encode reference semantics. To reduce this shortcut, we construct training tuples from MegaStyle-1M \cite{gao2026megastyle}. Each tuple contains reference images and a target image that share style but differ in content, and the prompt describes the target content rather than the references. The loss therefore rewards style consistency while discouraging object or identity copying as a shortcut. In practice, we sample multiple references when available to improve robustness, while retaining single-image examples to support one-shot inference.

\subsection{Asymmetric LoRA Guidance}

Because i2L converts LoRA training into inference-time weight prediction, producing an additional LoRA introduces little overhead. We exploit this property for asymmetric LoRA guidance. Let $\Delta\Theta_{\mathcal{R}}=G_{\phi}(\mathcal{R})$ denote the reference-style LoRA, and let $\Delta\Theta_{\varnothing}=G_{\phi}(r_{\mathrm{gray}})$ denote a neutral LoRA predicted from a pure gray image. Classifier-free guidance combines two predictions \cite{ho2022classifierfree}:
\begin{align}
    v_{\mathrm{neg}} &= v_{\theta+\Delta\Theta_{\varnothing}}(x_t,t,c_{\varnothing}), \\ 
    v_{\mathrm{pos}} &= v_{\theta+\Delta\Theta_{\mathcal{R}}}(x_t,t,c), \\ 
    \hat{v} &= v_{\mathrm{neg}} + s\left(v_{\mathrm{pos}}-v_{\mathrm{neg}}\right).
\end{align}
where $s$ is the guidance scale and $c_{\varnothing}$ denotes the negative or empty text condition. Instead of sharing weights across the two branches, we apply the reference-image LoRA to the positive branch and the neutral gray-image LoRA to the negative branch. The gray-image LoRA serves as a style-neutral baseline, so the guidance direction emphasizes the visual characteristics introduced by the reference LoRA. This improves style adherence without additional optimization or sampler modifications.

\section{Experiments}

\begin{figure*}[htbp]
    \centering
    \setlength{\tabcolsep}{1pt}%
    \begin{tabular}{cccc}
        \begin{minipage}[b]{0.23\textwidth}\centering
            \begin{tabular}{@{}cc@{}}
                \includegraphics[width=0.48\linewidth,height=0.48\linewidth,keepaspectratio]{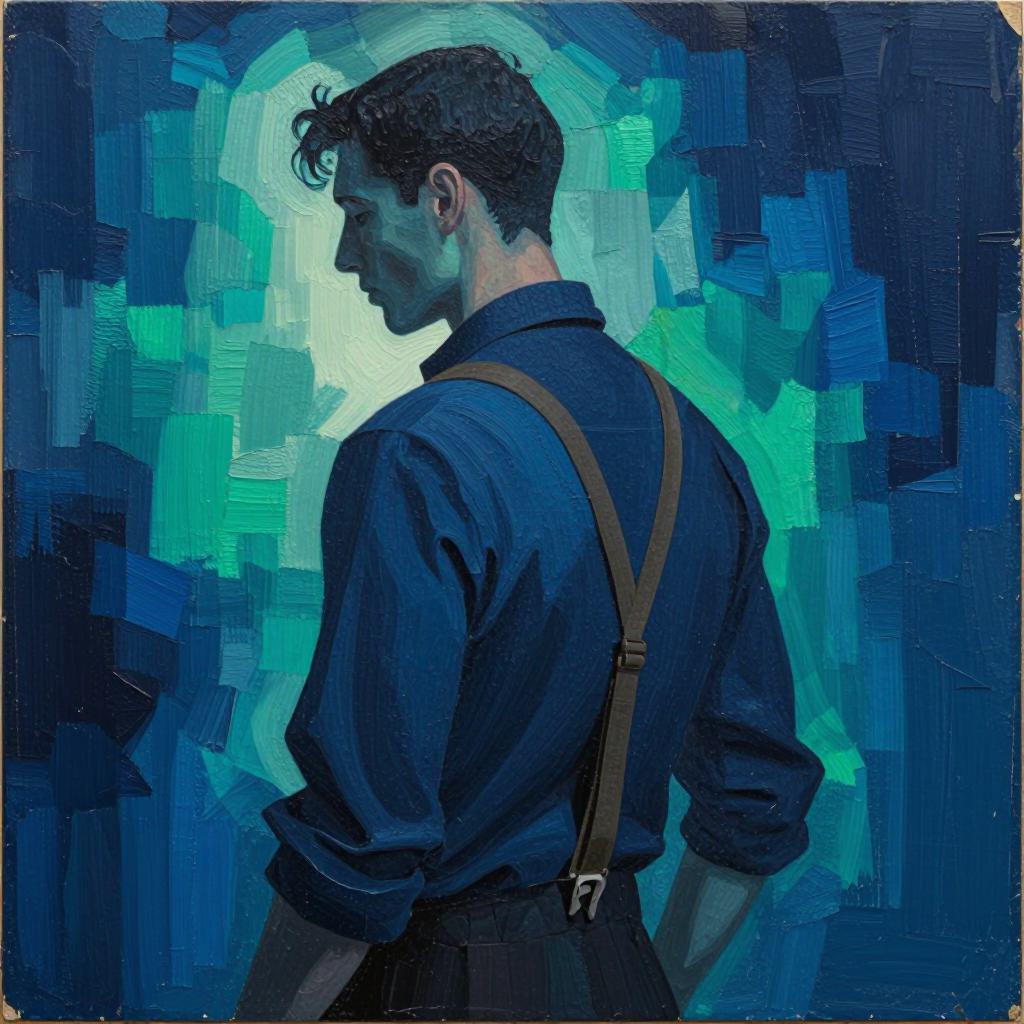} &
                \includegraphics[width=0.48\linewidth,height=0.48\linewidth,keepaspectratio]{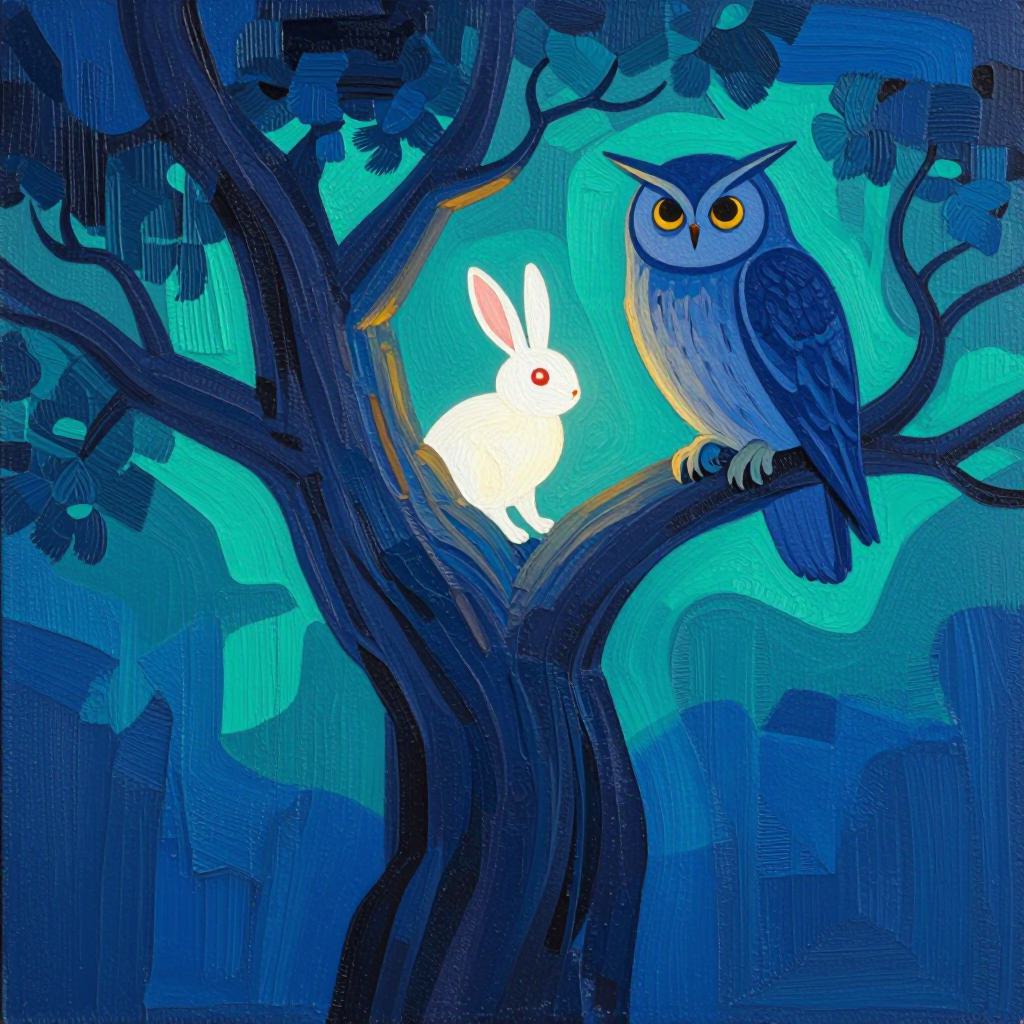} \\
                \includegraphics[width=0.48\linewidth,height=0.48\linewidth,keepaspectratio]{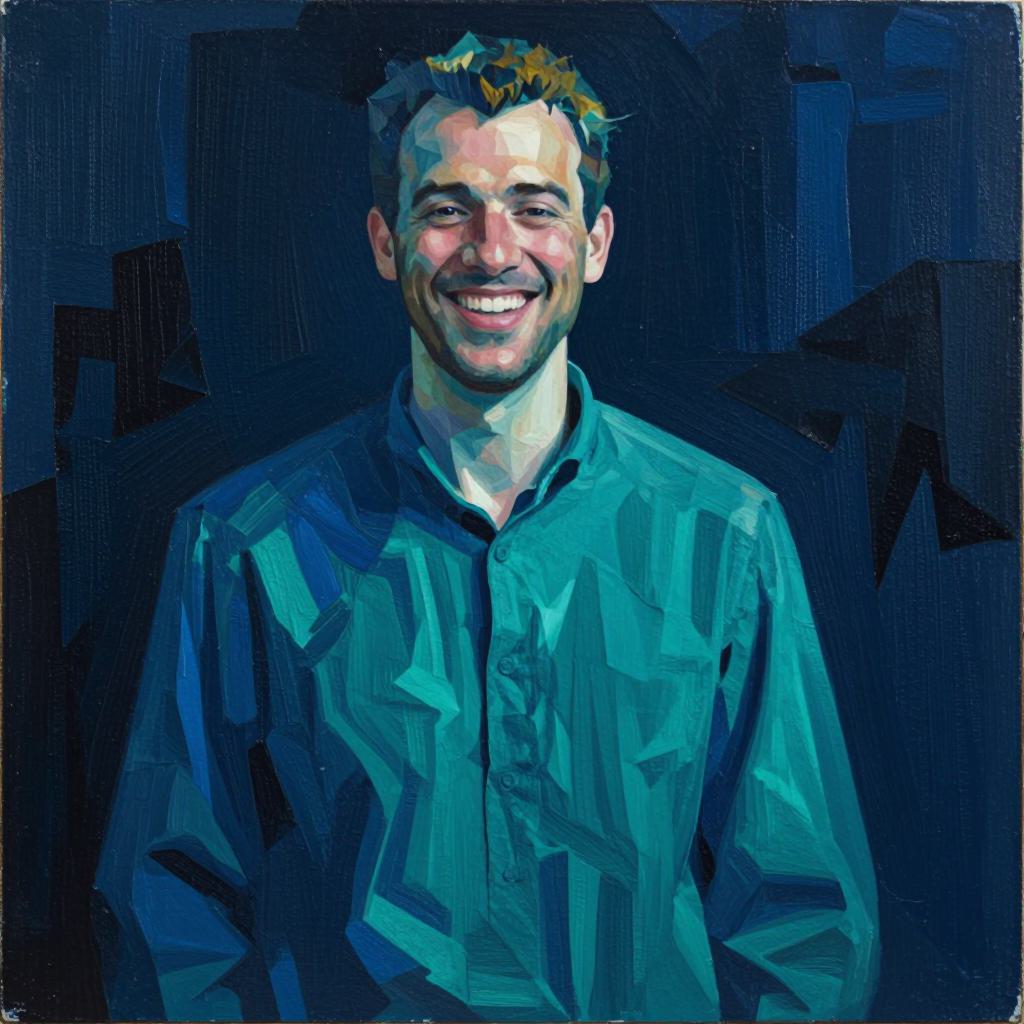} &
                \includegraphics[width=0.48\linewidth,height=0.48\linewidth,keepaspectratio]{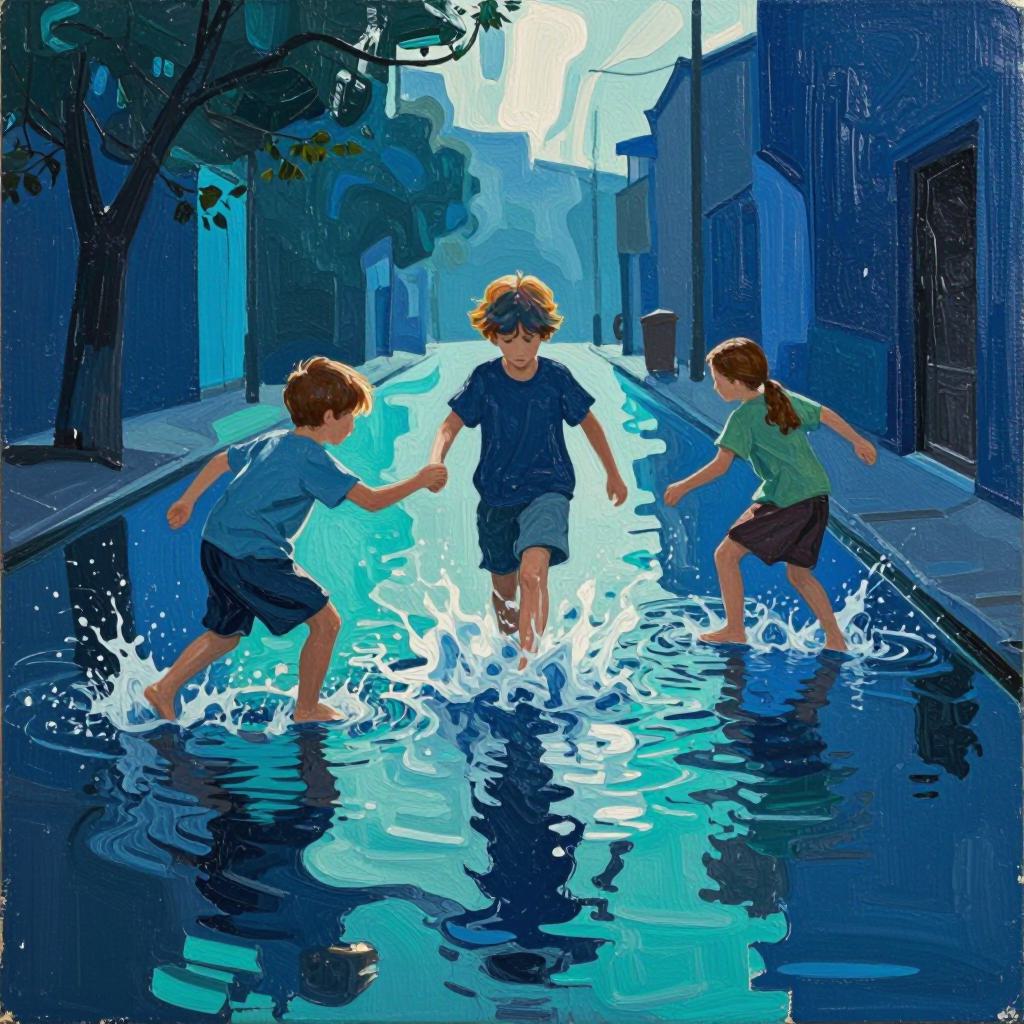}
            \end{tabular}\\
            {\scriptsize Inputs A--D}
        \end{minipage} &
        \imgcellb{0.23\textwidth}{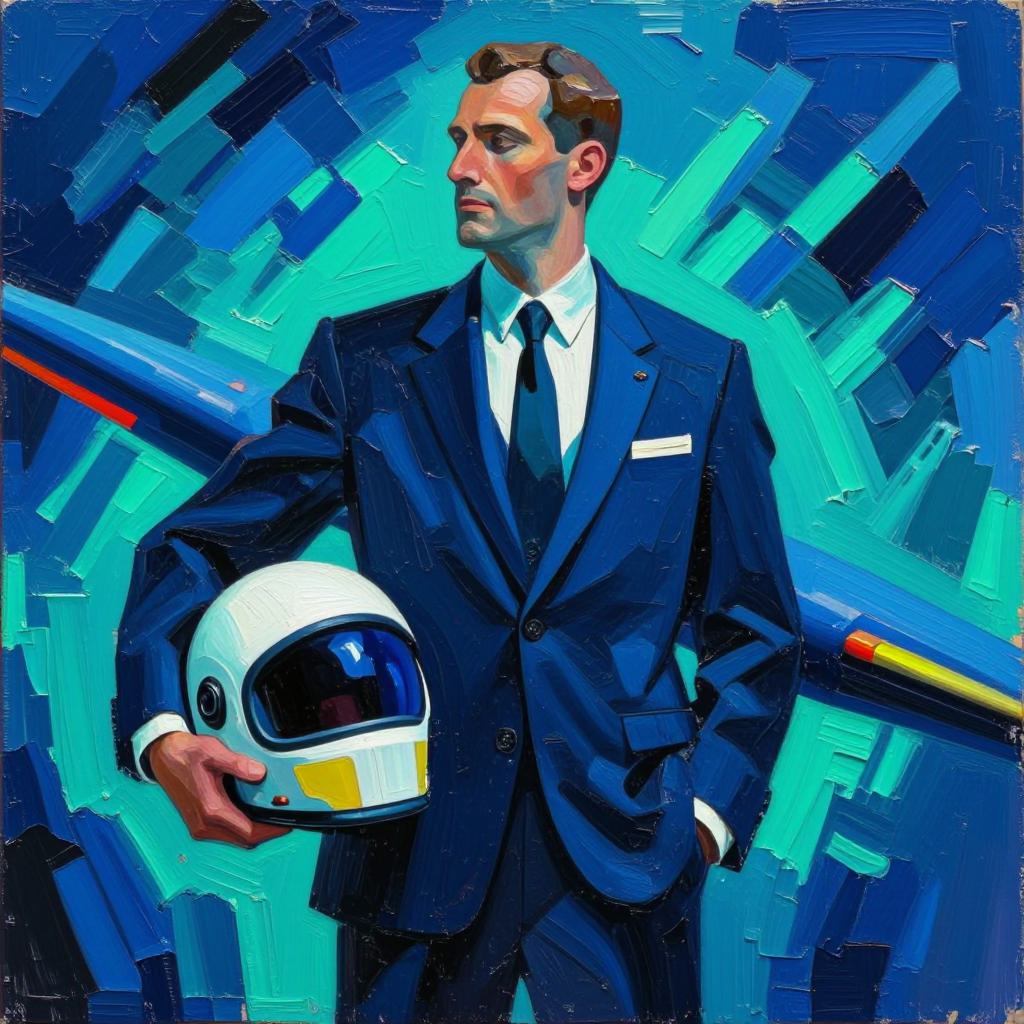}{i2L-Z-Image} &
        \imgcellb{0.23\textwidth}{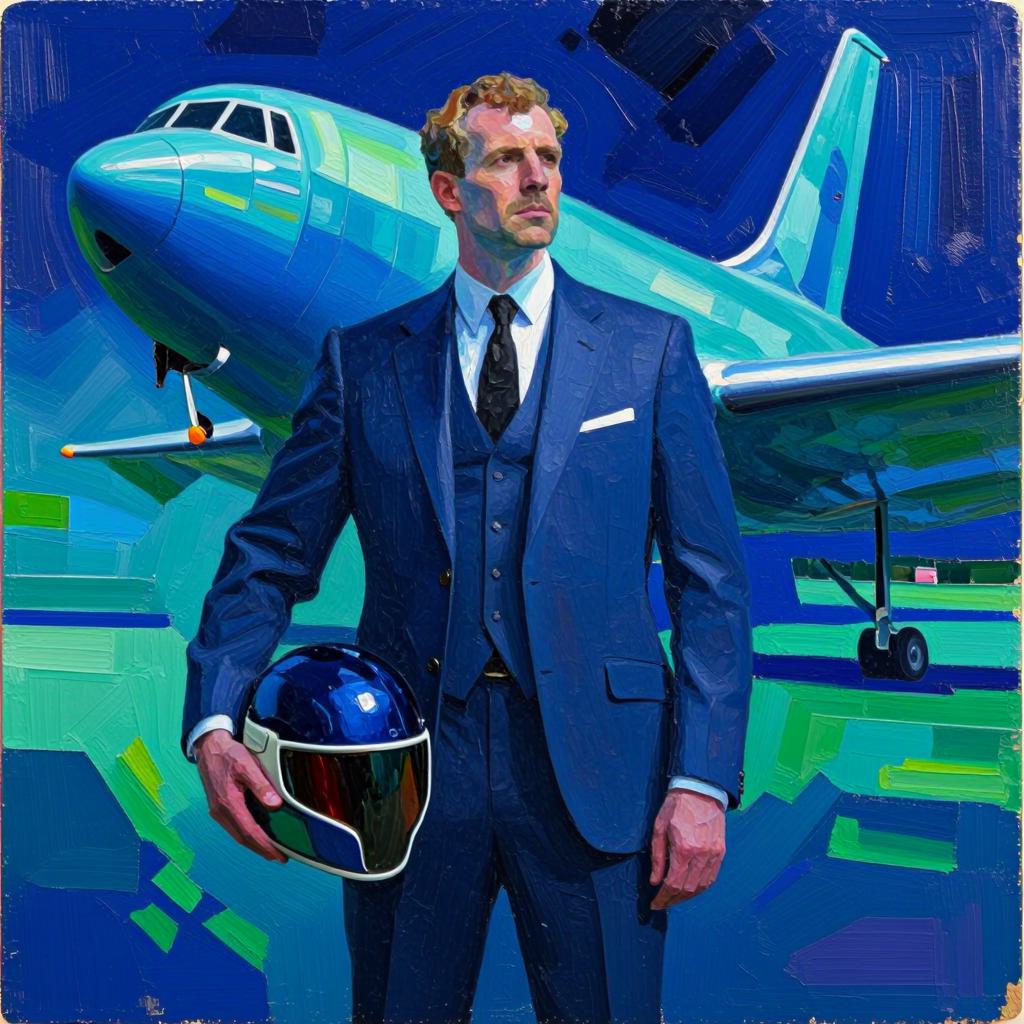}{i2L-FLUX.2} &
        \imgcellb{0.23\textwidth}{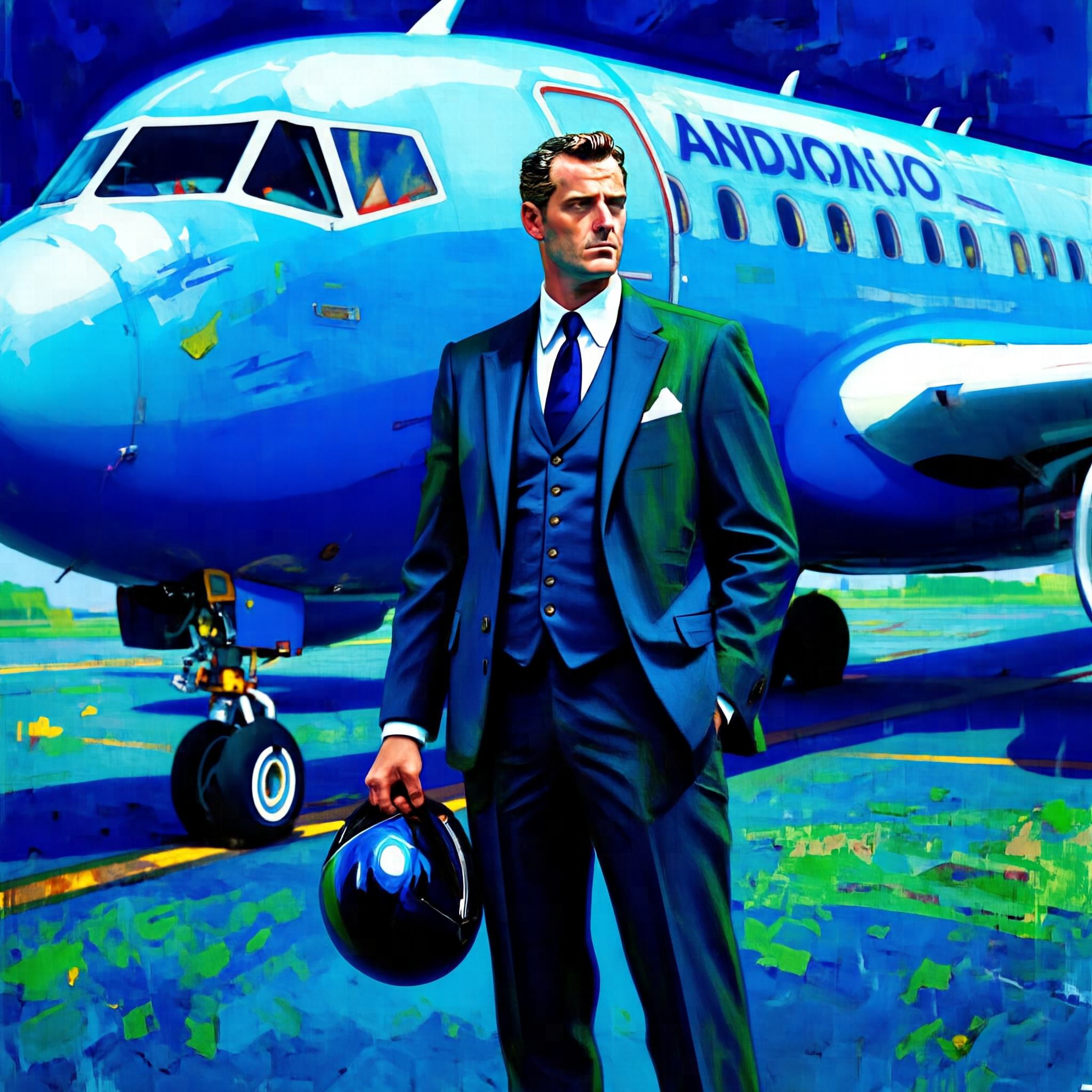}{i2L-Hidream-O1}
    \end{tabular}

    \vspace{0.15em}
    \begin{tabular}{cccccccc}
        \imgcell{0.112\textwidth}{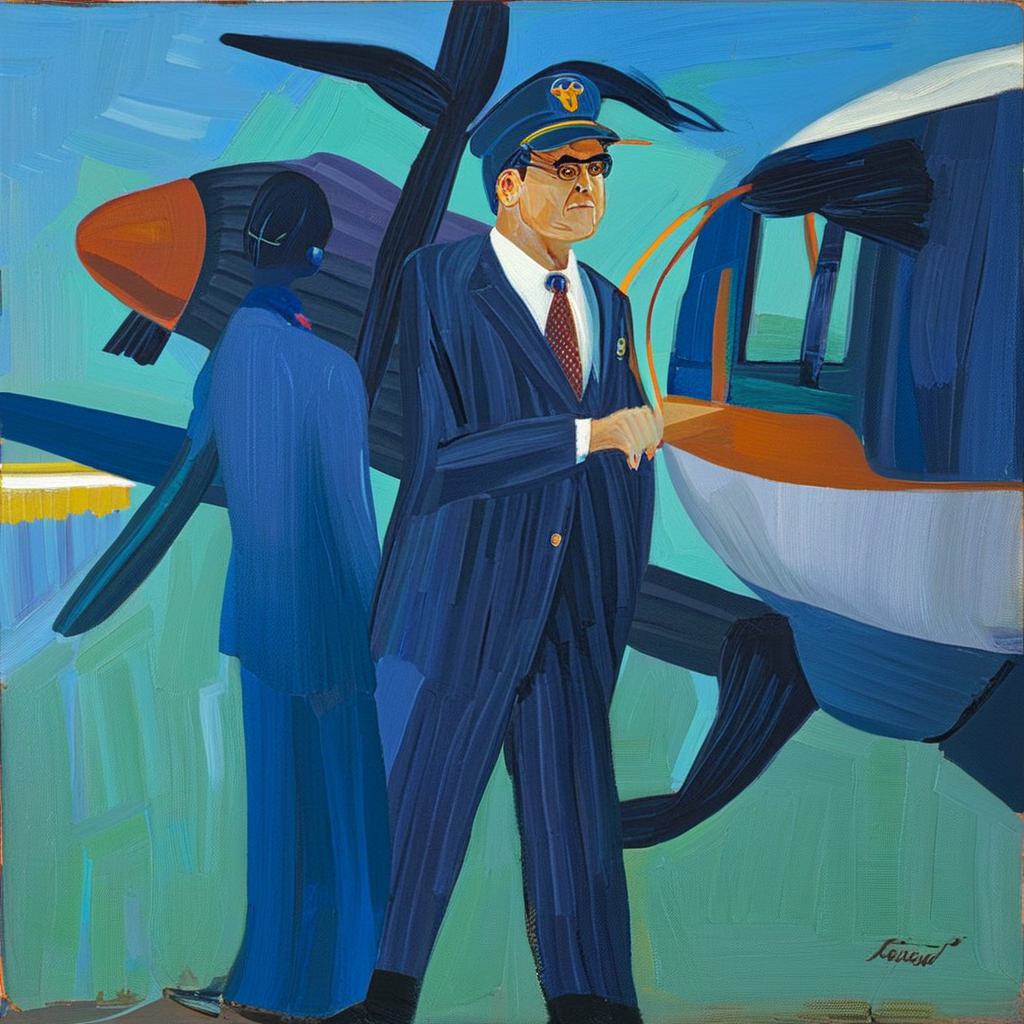}{StyleCrafter} &
        \imgcell{0.112\textwidth}{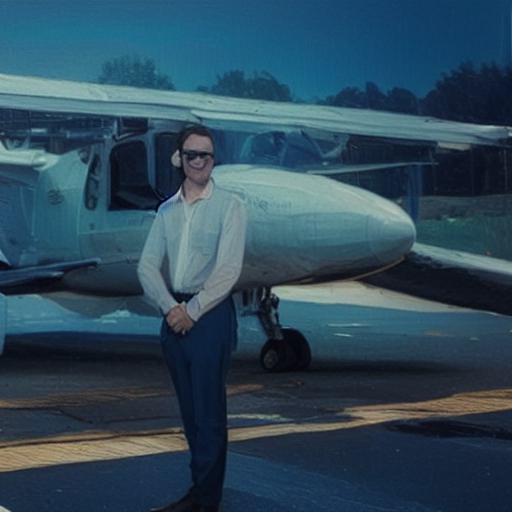}{StyleID} &
        \imgcell{0.112\textwidth}{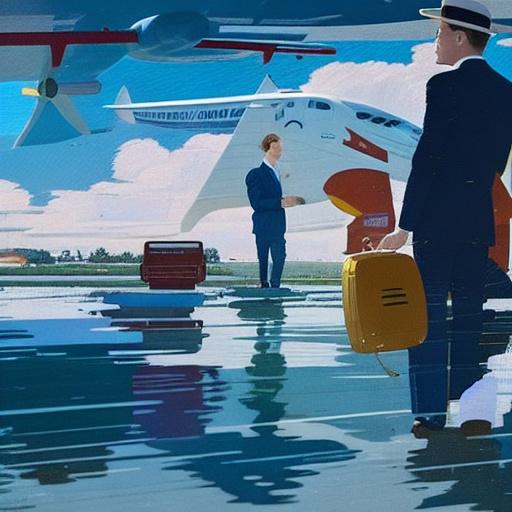}{ControlNet} &
        \imgcell{0.112\textwidth}{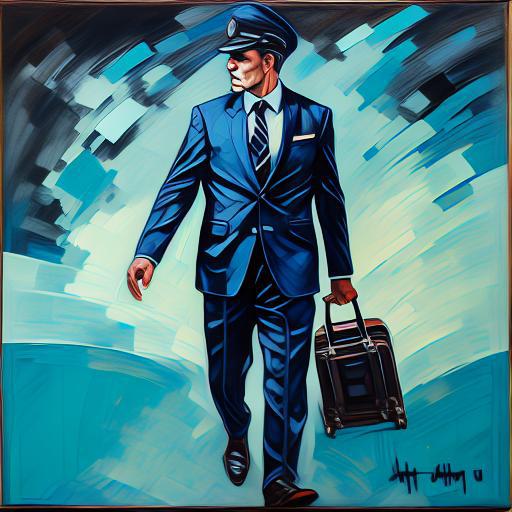}{DEADiff} &
        \imgcell{0.112\textwidth}{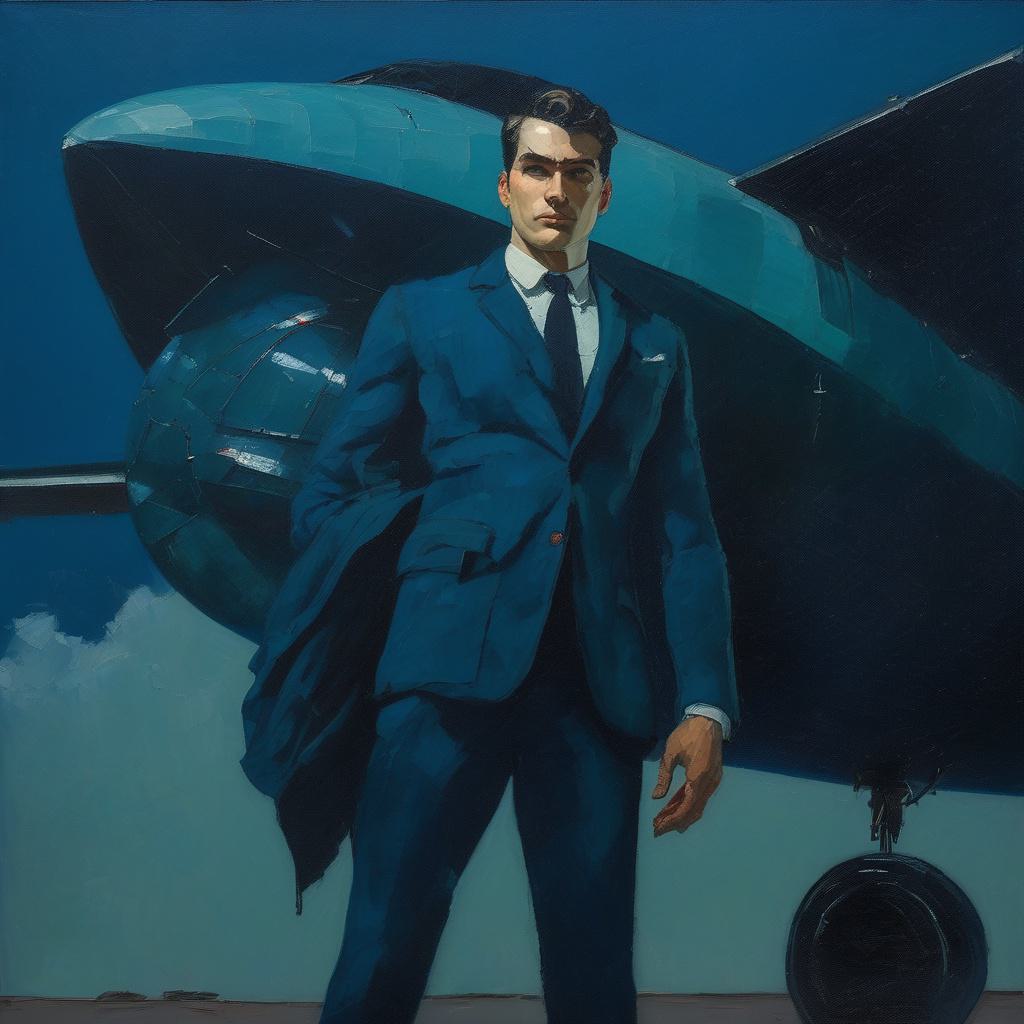}{InstantStyle} &
        \imgcell{0.112\textwidth}{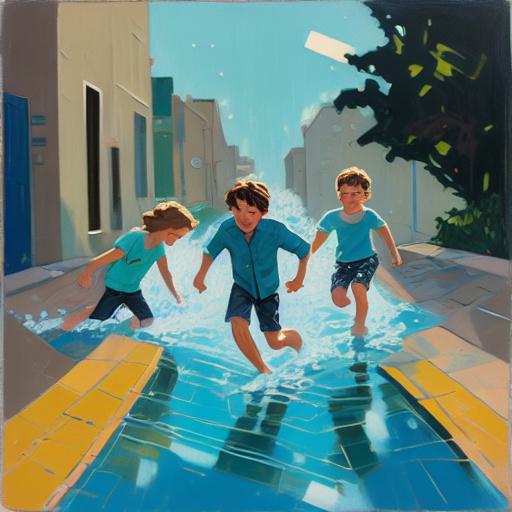}{IP-Adapter} &
        \imgcell{0.112\textwidth}{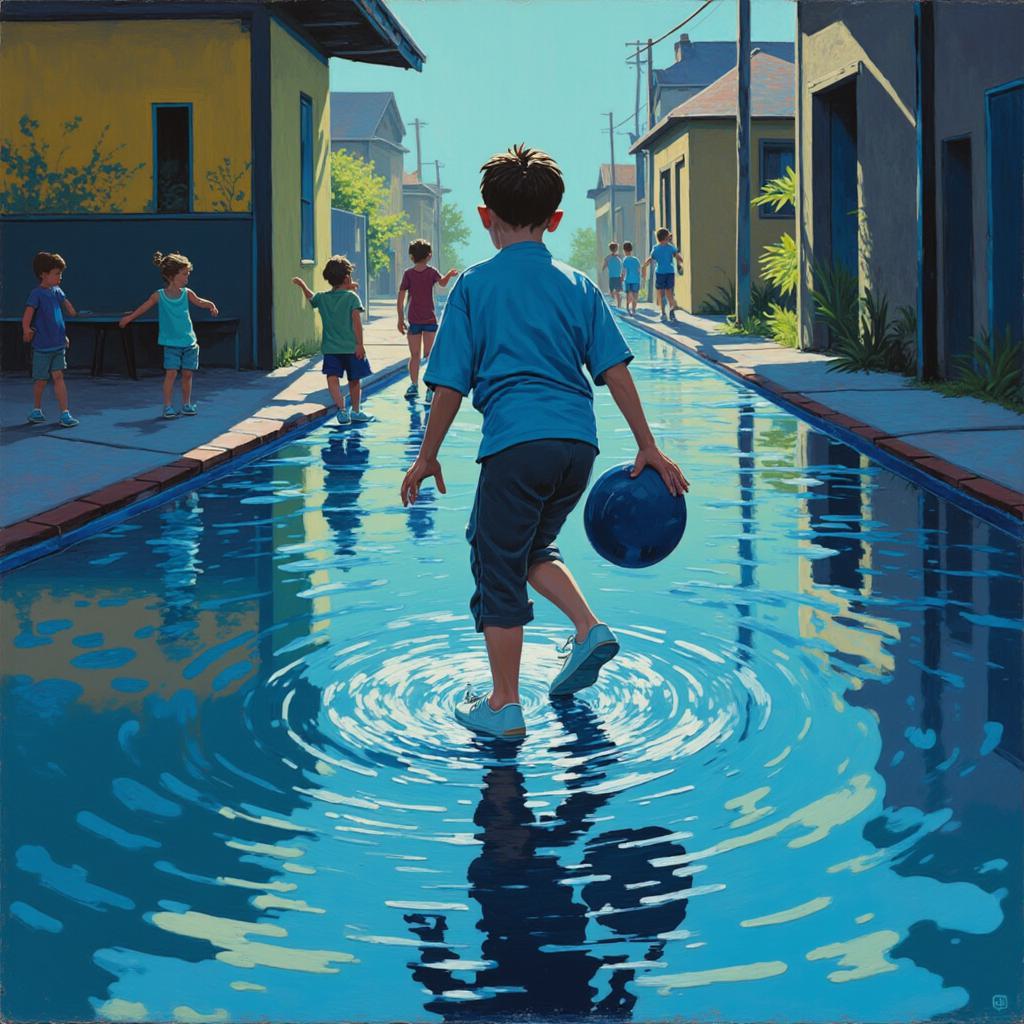}{IP-Adapter-FLUX} &
        \imgcell{0.112\textwidth}{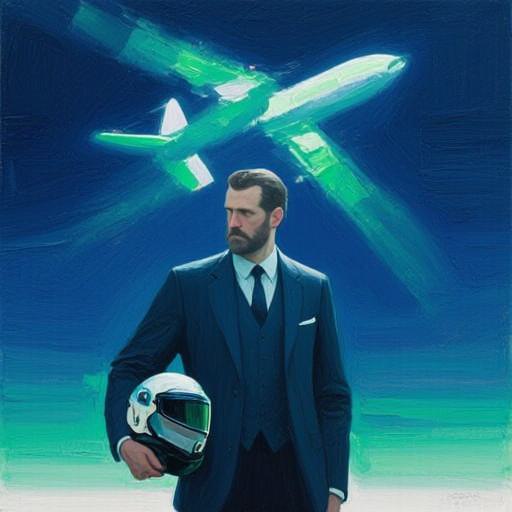}{MegaStyle-FLUX}
    \end{tabular}%

    \vspace{0.5em}
    \setlength{\tabcolsep}{1pt}%
    \begin{tabular}{cccc}
        \begin{minipage}[b]{0.23\textwidth}\centering
            \begin{tabular}{@{}cc@{}}
                \includegraphics[width=0.48\linewidth,height=0.48\linewidth,keepaspectratio]{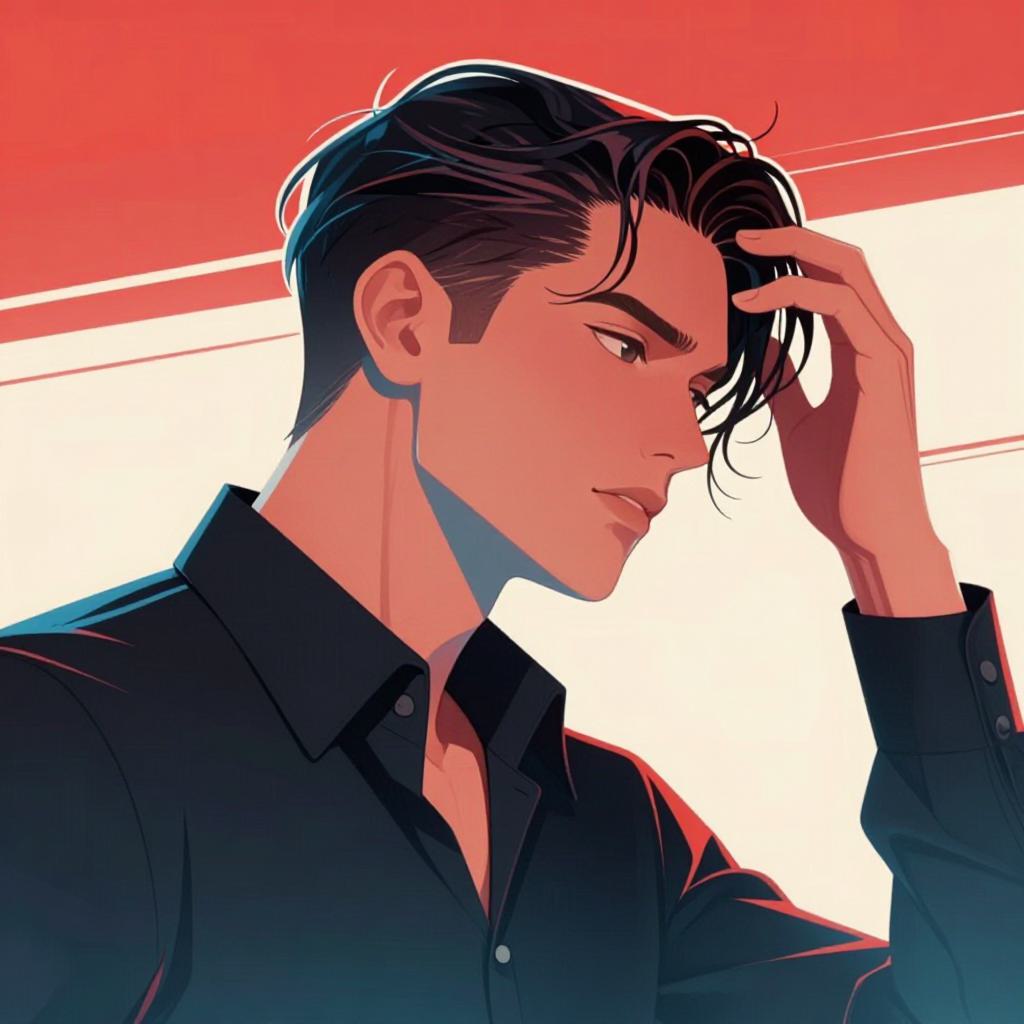} &
                \includegraphics[width=0.48\linewidth,height=0.48\linewidth,keepaspectratio]{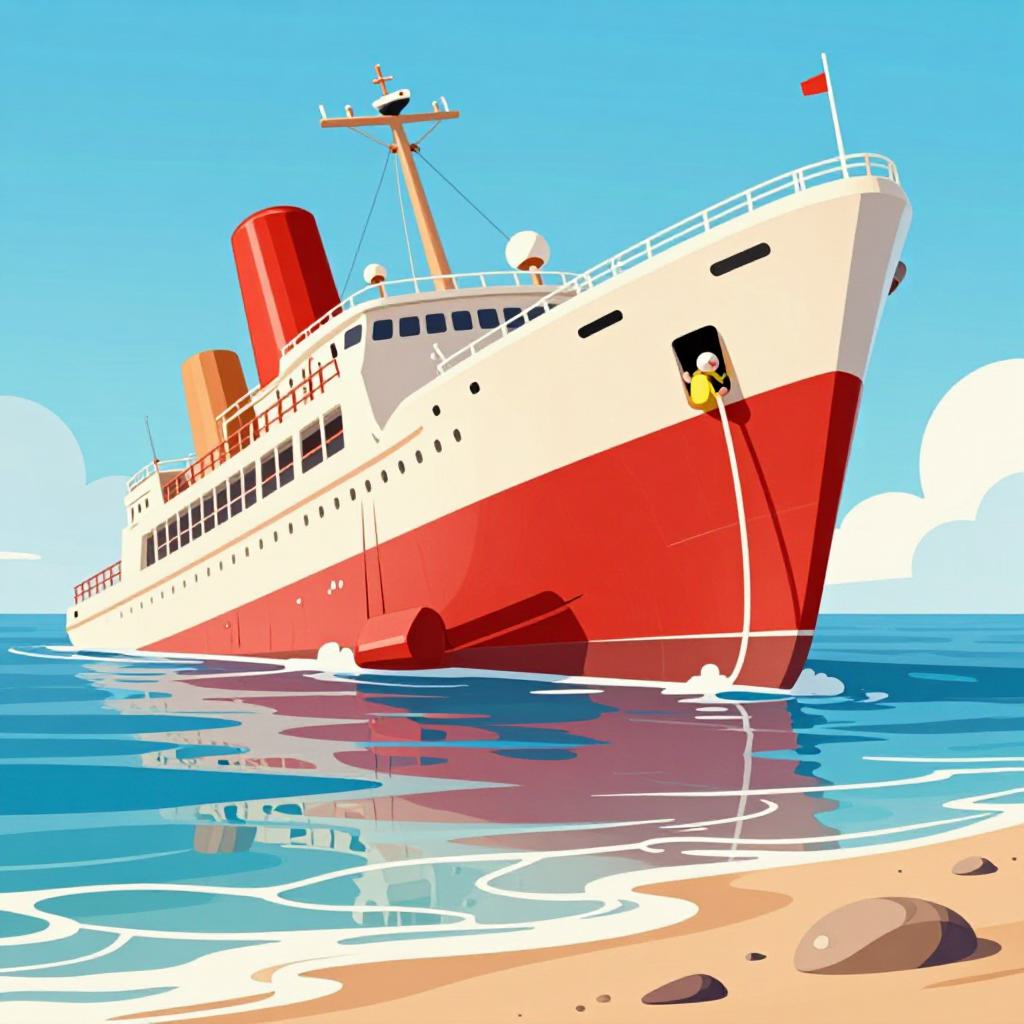} \\
                \includegraphics[width=0.48\linewidth,height=0.48\linewidth,keepaspectratio]{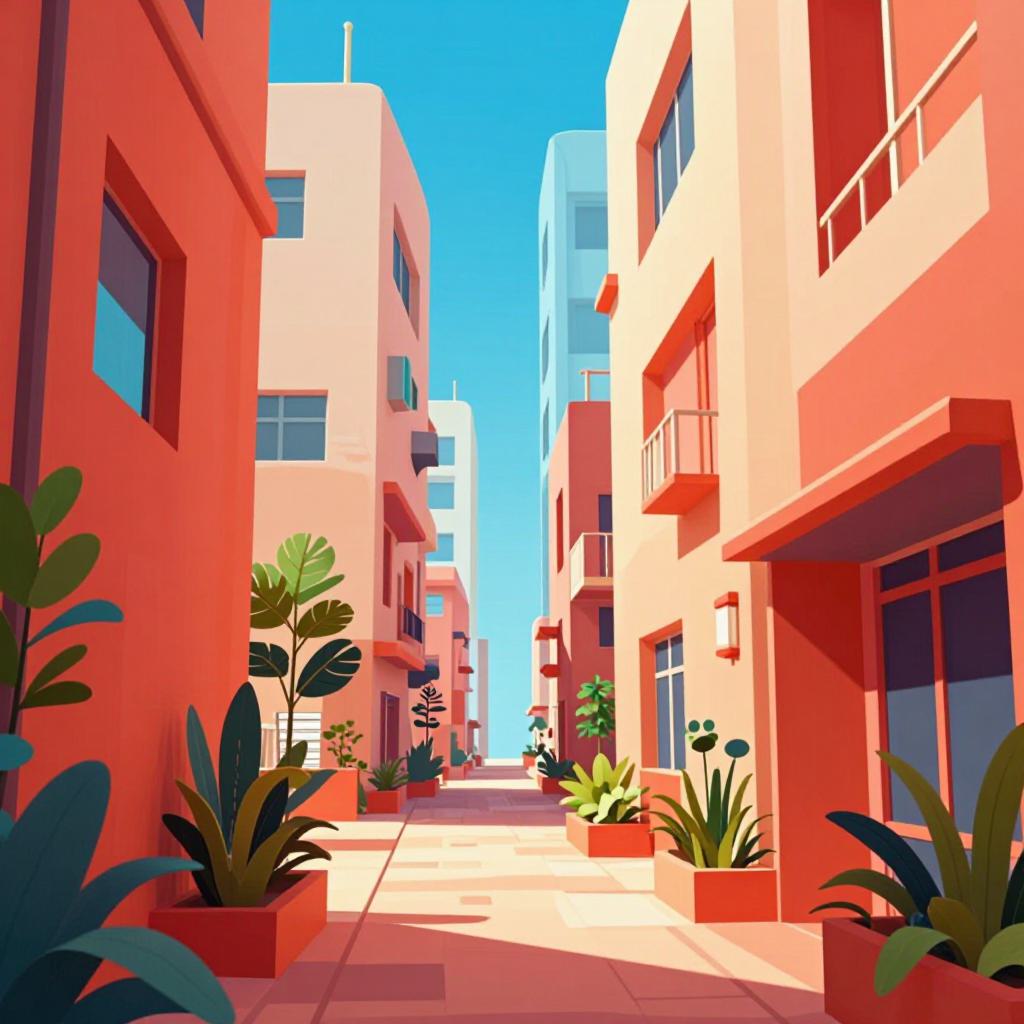} &
                \includegraphics[width=0.48\linewidth,height=0.48\linewidth,keepaspectratio]{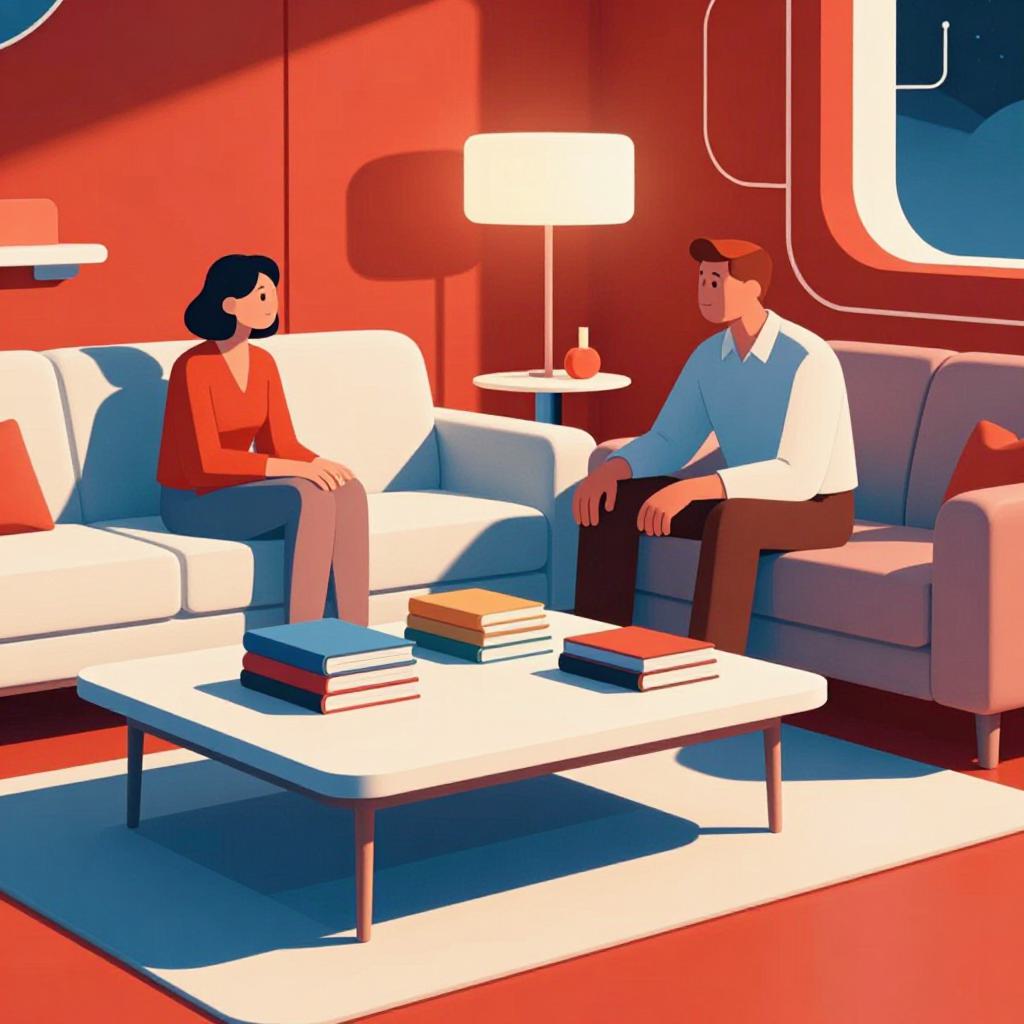}
            \end{tabular}\\
            {\scriptsize Inputs A--D}
        \end{minipage} &
        \imgcellb{0.23\textwidth}{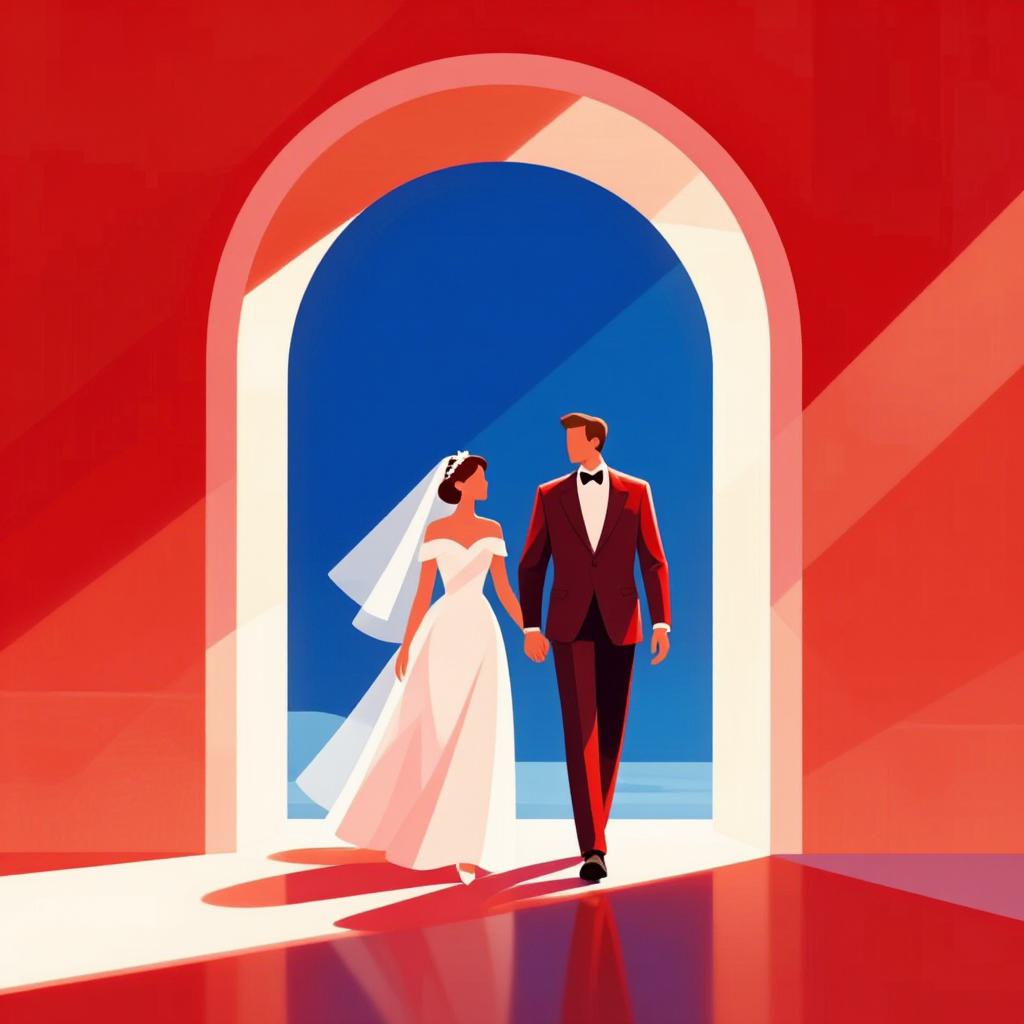}{i2L-Z-Image} &
        \imgcellb{0.23\textwidth}{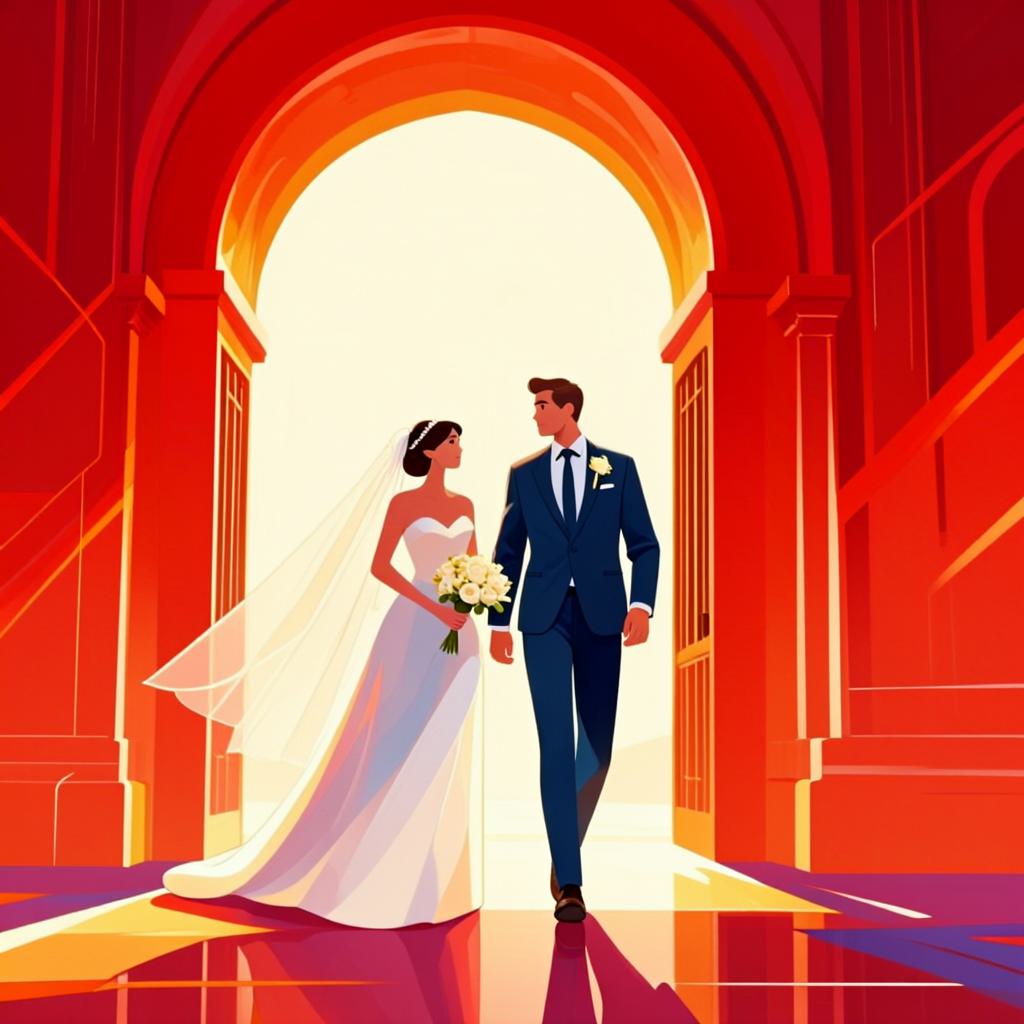}{i2L-FLUX.2} &
        \imgcellb{0.23\textwidth}{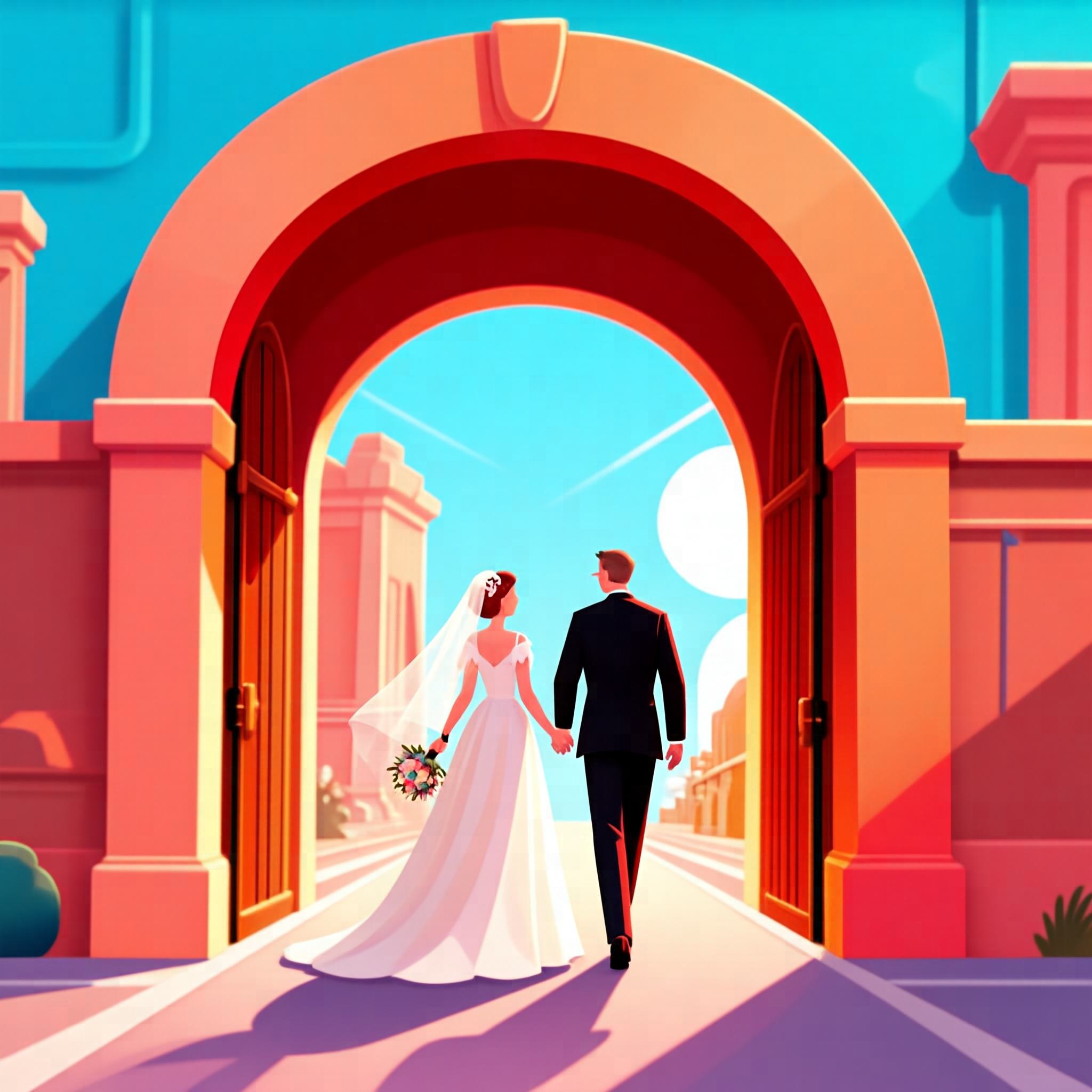}{i2L-Hidream-O1}
    \end{tabular}

    \vspace{0.15em}
    \begin{tabular}{cccccccc}
        \imgcell{0.112\textwidth}{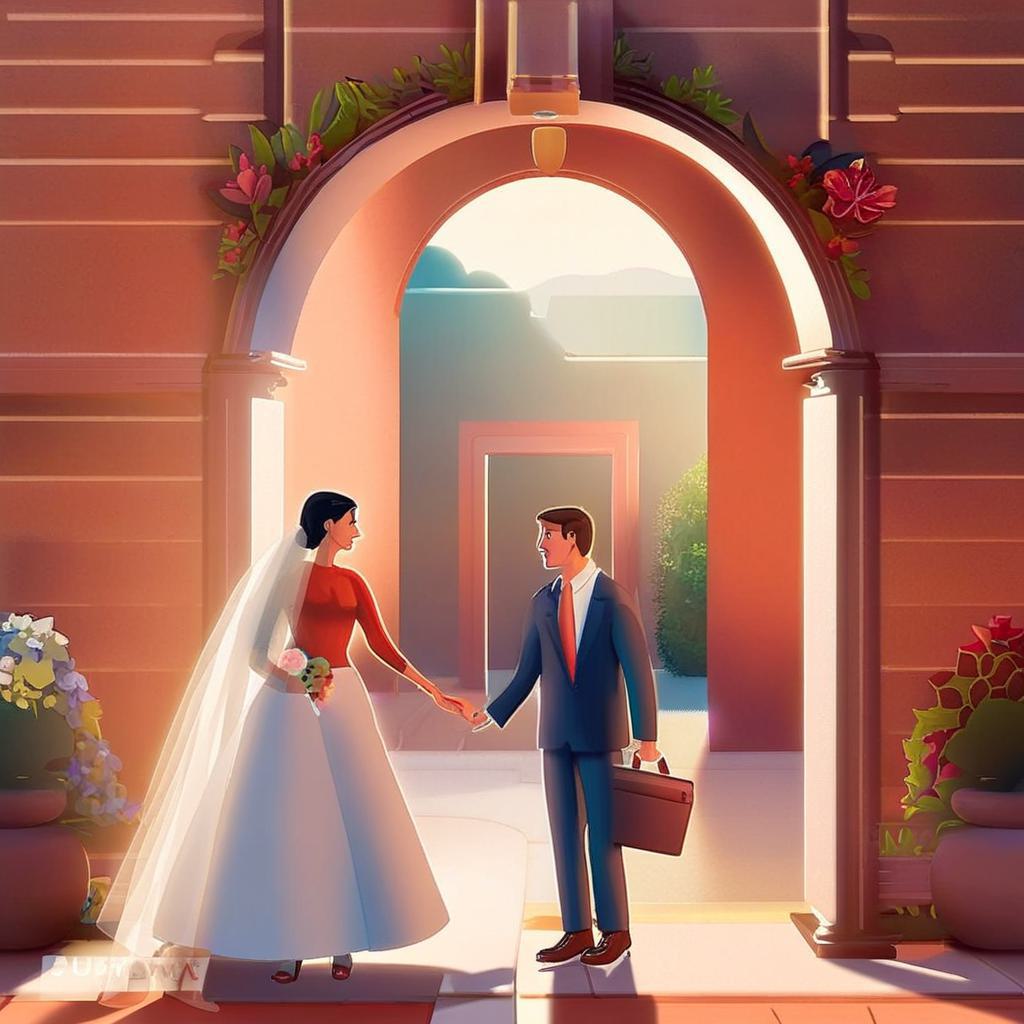}{StyleCrafter} &
        \imgcell{0.112\textwidth}{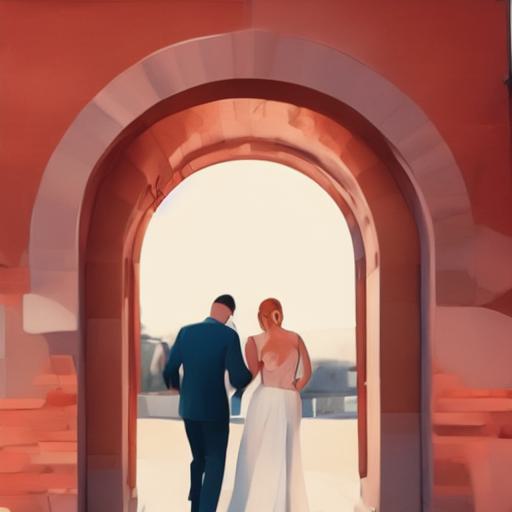}{StyleID} &
        \imgcell{0.112\textwidth}{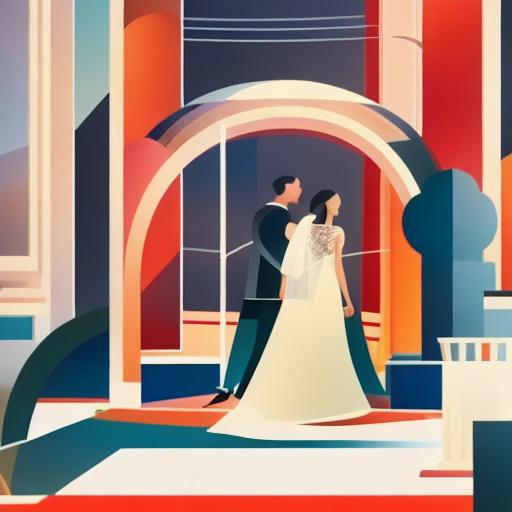}{ControlNet} &
        \imgcell{0.112\textwidth}{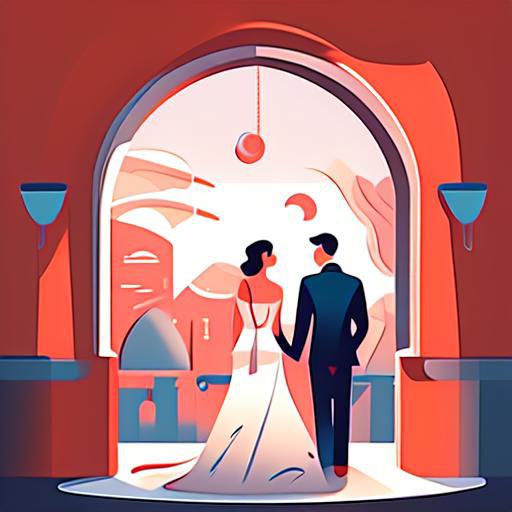}{DEADiff} &
        \imgcell{0.112\textwidth}{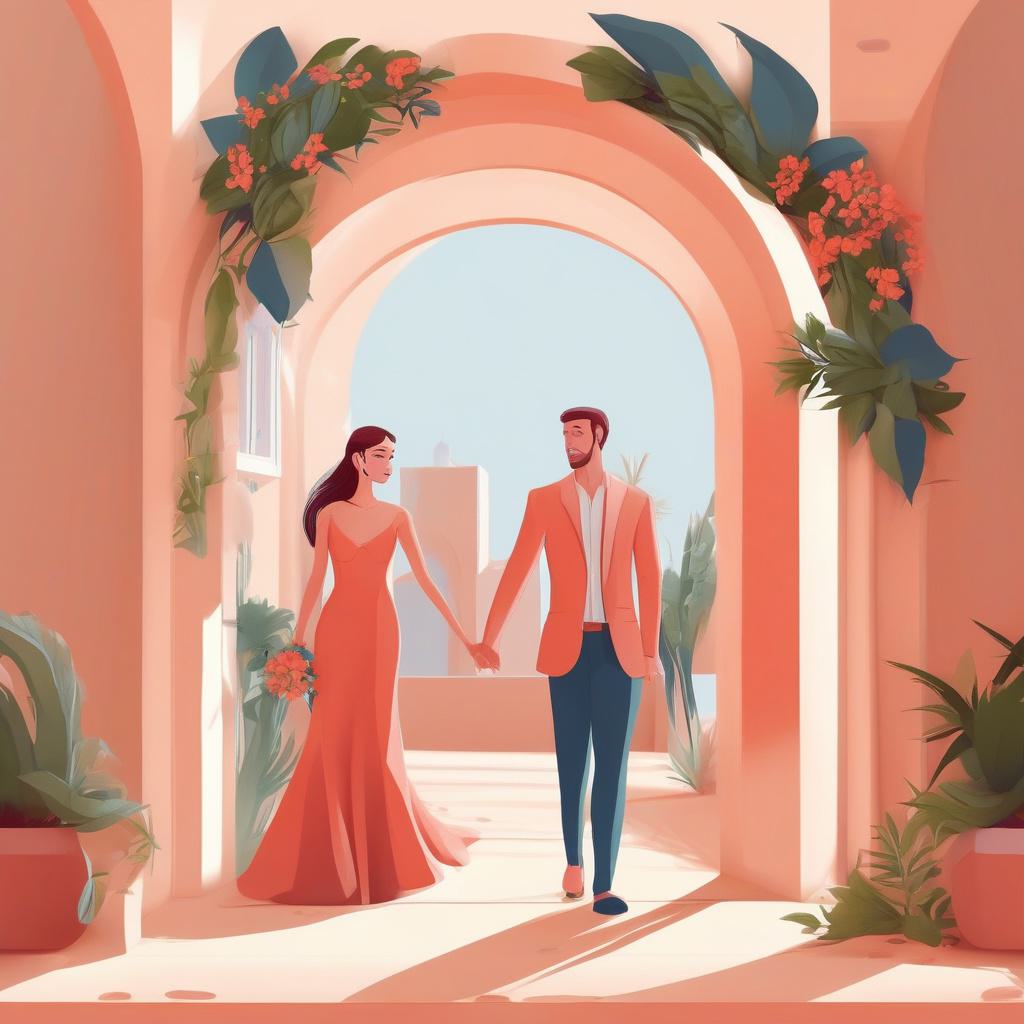}{InstantStyle} &
        \imgcell{0.112\textwidth}{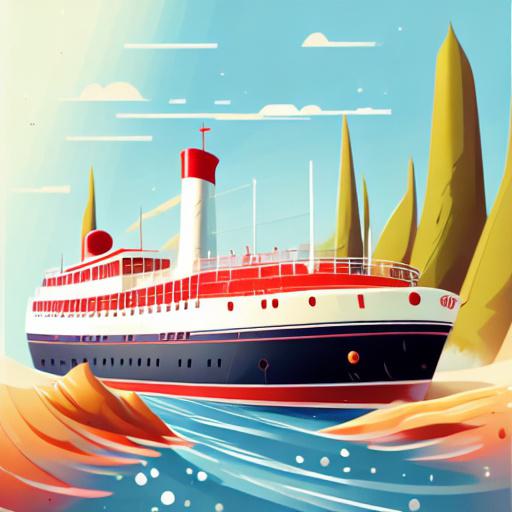}{IP-Adapter} &
        \imgcell{0.112\textwidth}{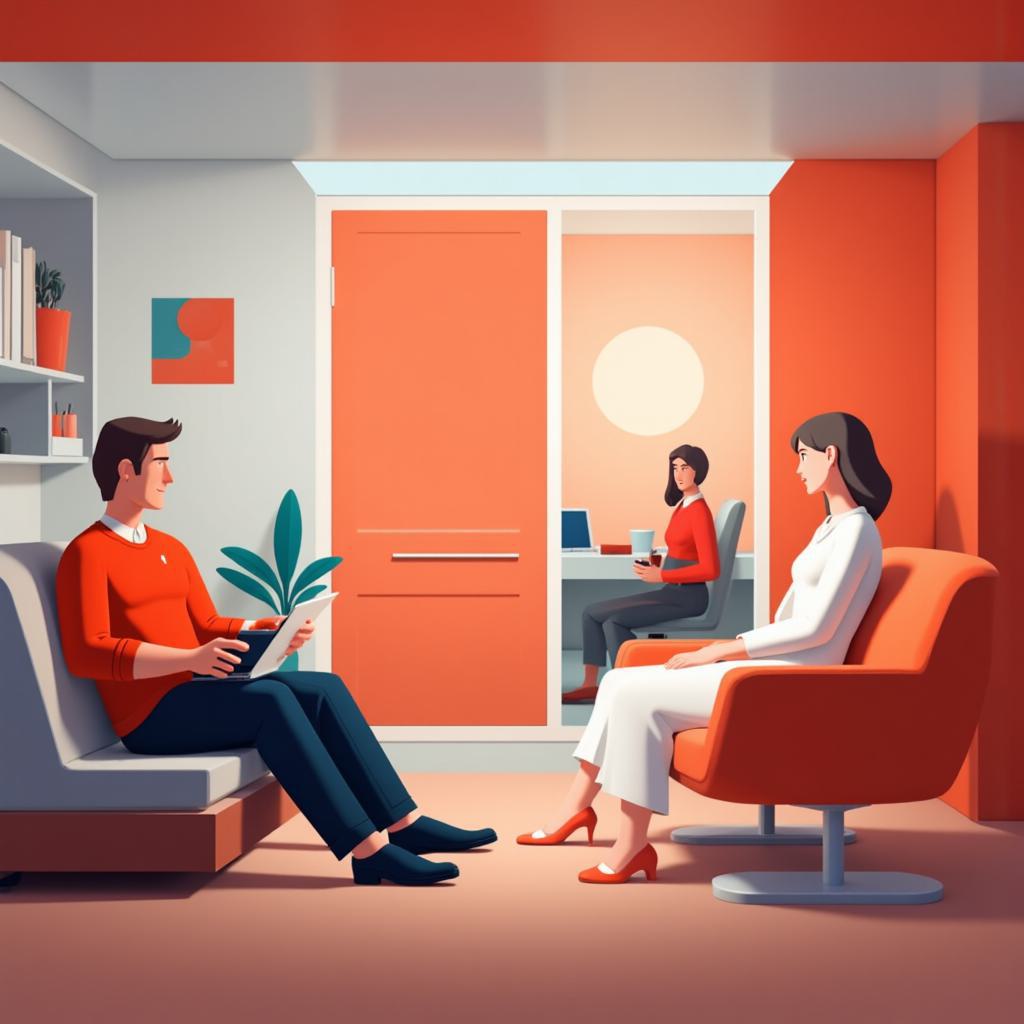}{IP-Adapter-FLUX} &
        \imgcell{0.112\textwidth}{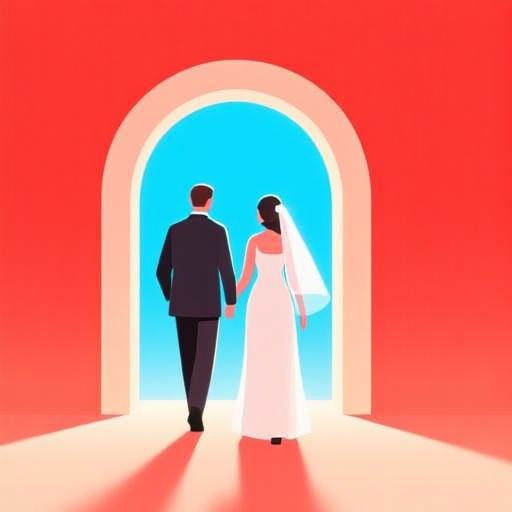}{MegaStyle-FLUX}
    \end{tabular}%

    \vspace{0.5em}
    \setlength{\tabcolsep}{1pt}%
    \begin{tabular}{cccc}
        \begin{minipage}[b]{0.23\textwidth}\centering
            \begin{tabular}{@{}cc@{}}
                \includegraphics[width=0.48\linewidth,height=0.48\linewidth,keepaspectratio]{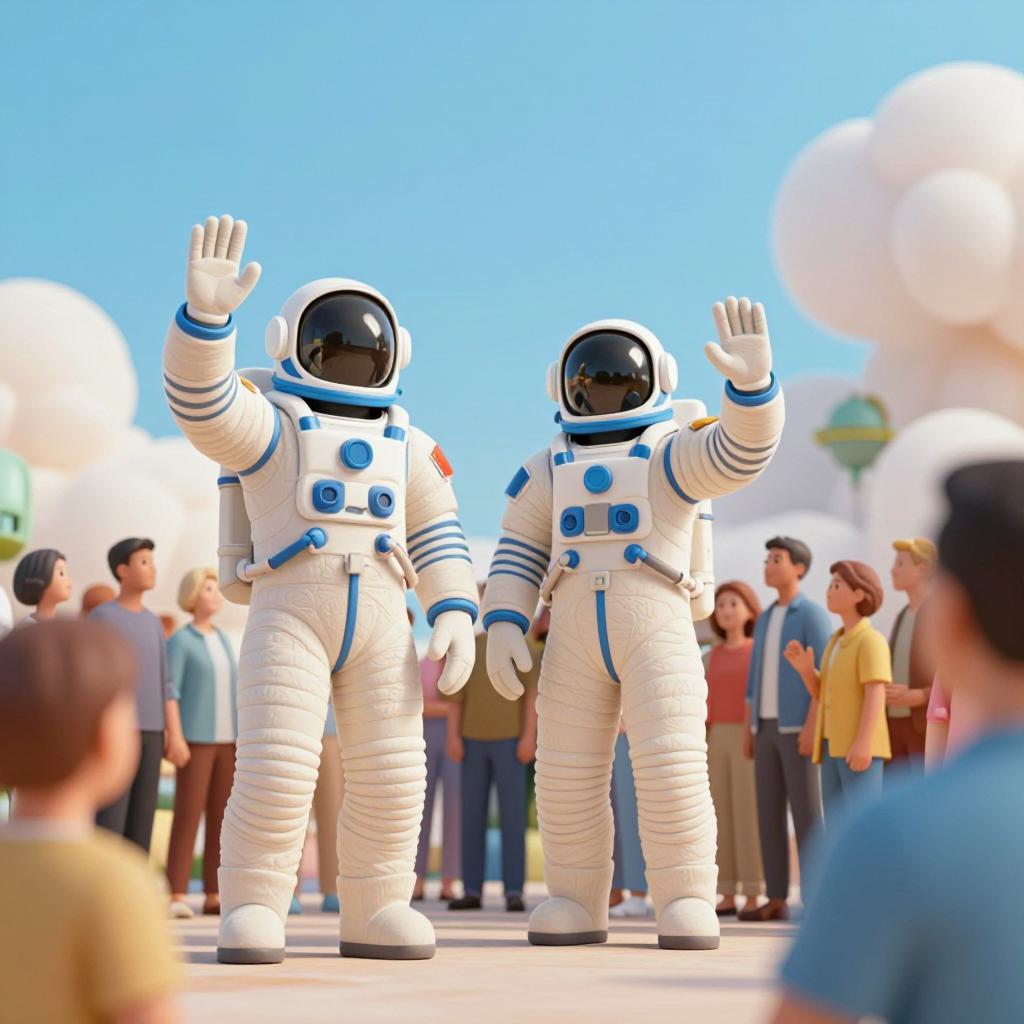} &
                \includegraphics[width=0.48\linewidth,height=0.48\linewidth,keepaspectratio]{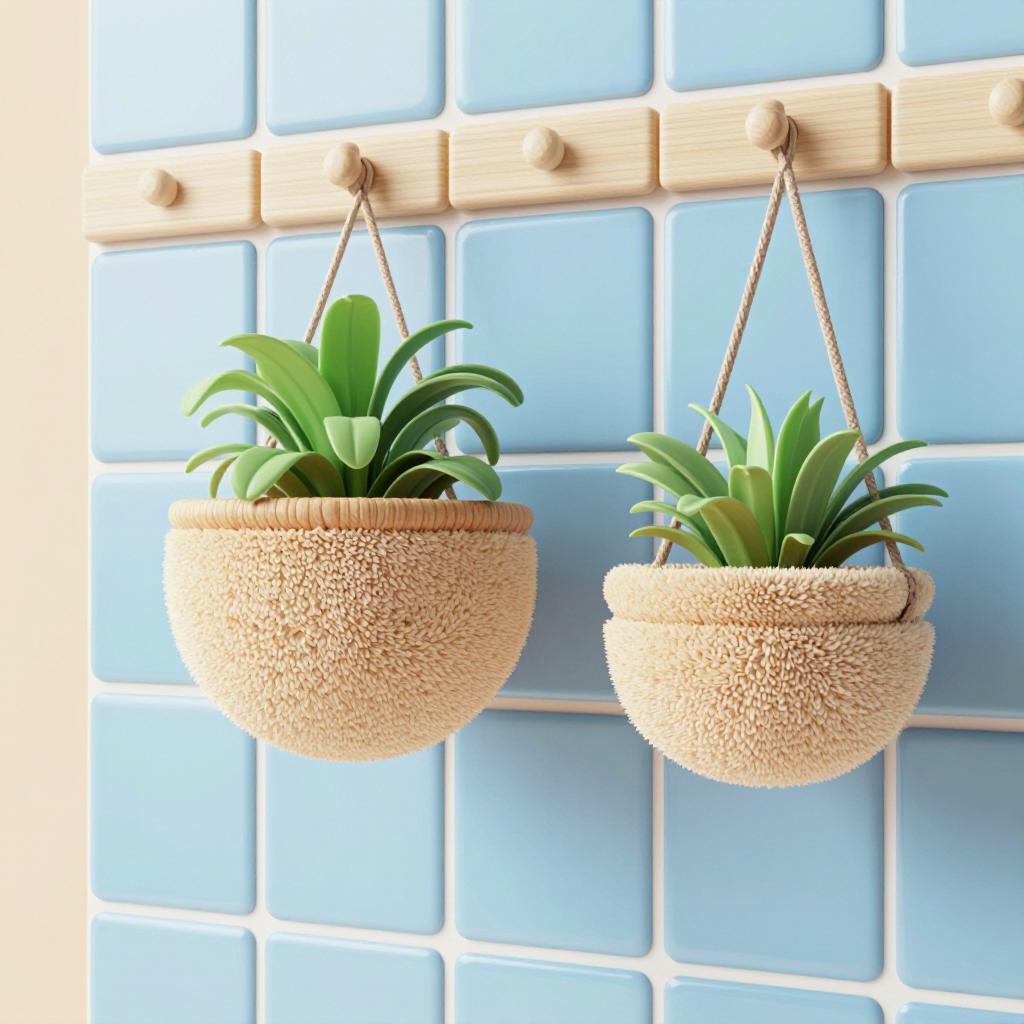} \\
                \includegraphics[width=0.48\linewidth,height=0.48\linewidth,keepaspectratio]{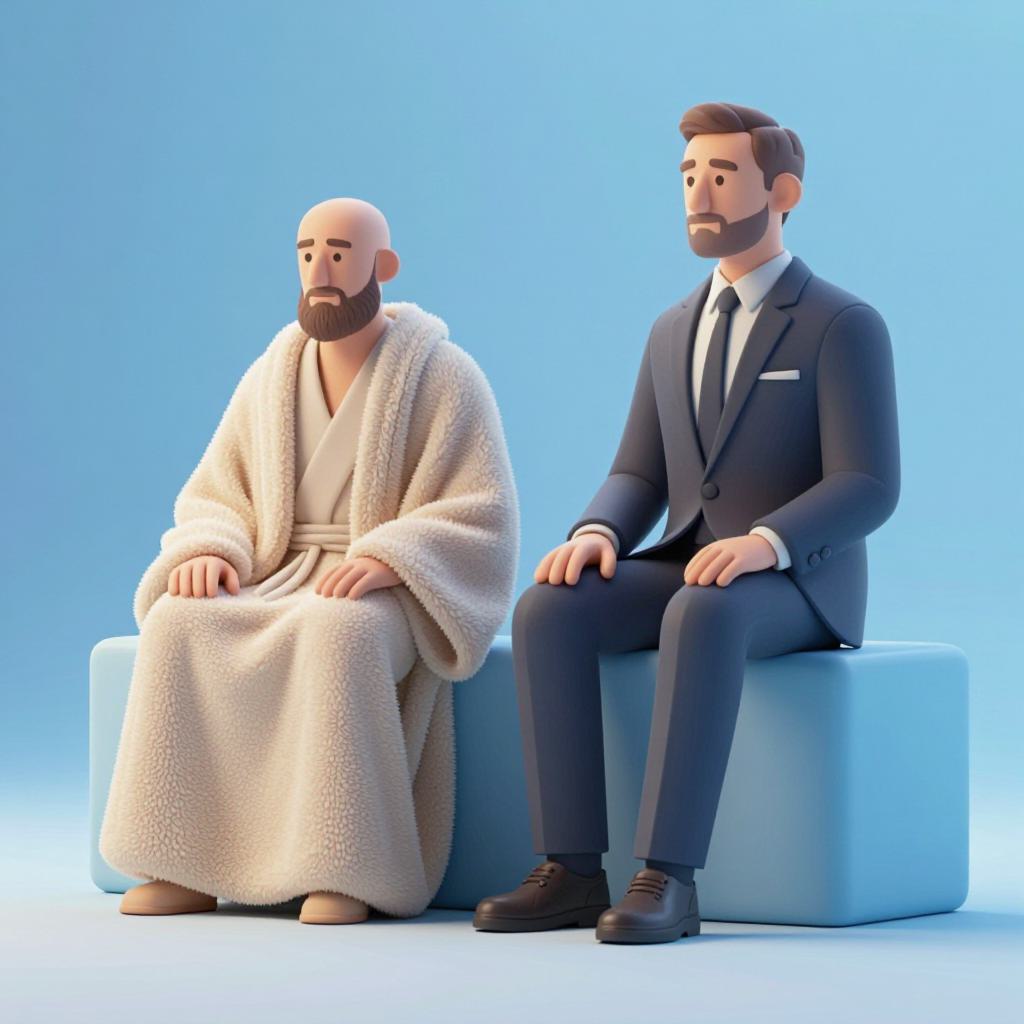} &
                \includegraphics[width=0.48\linewidth,height=0.48\linewidth,keepaspectratio]{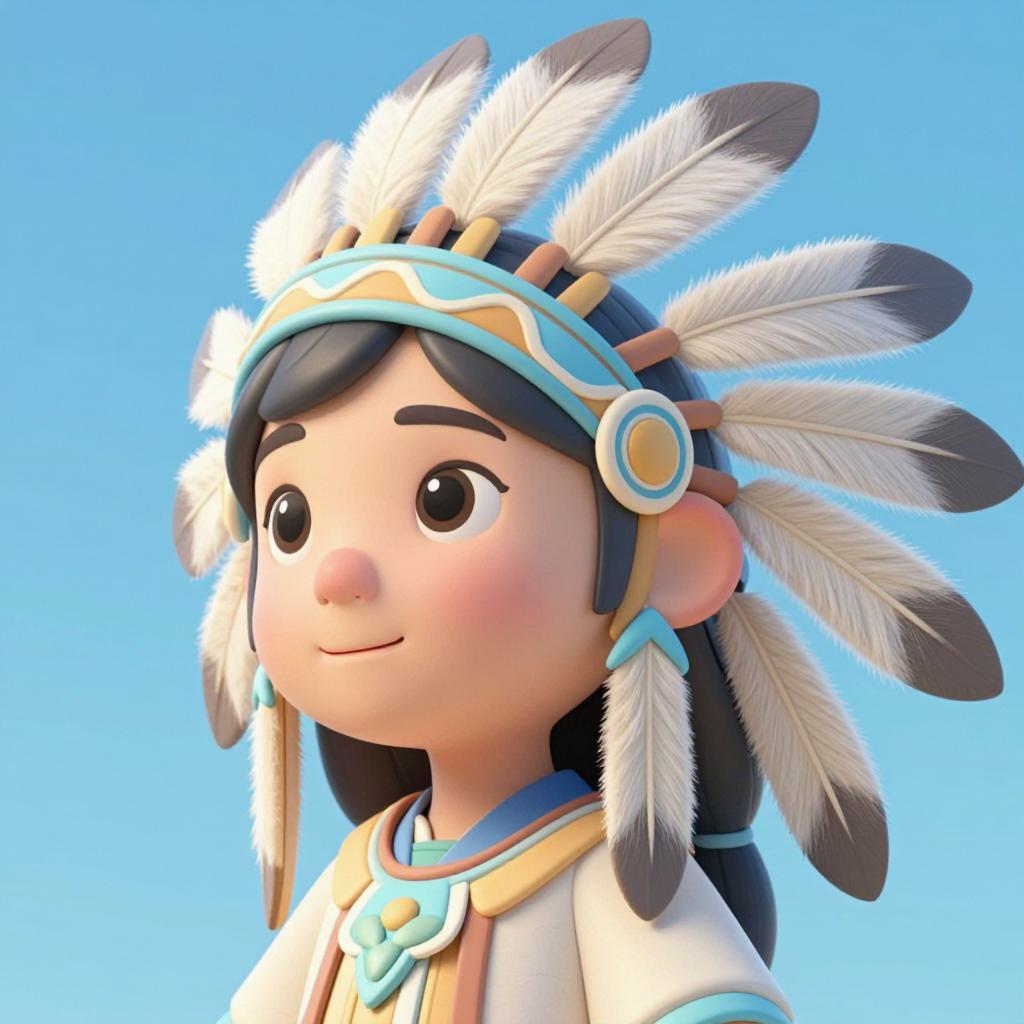}
            \end{tabular}\\
            {\scriptsize Inputs A--D}
        \end{minipage} &
        \imgcellb{0.23\textwidth}{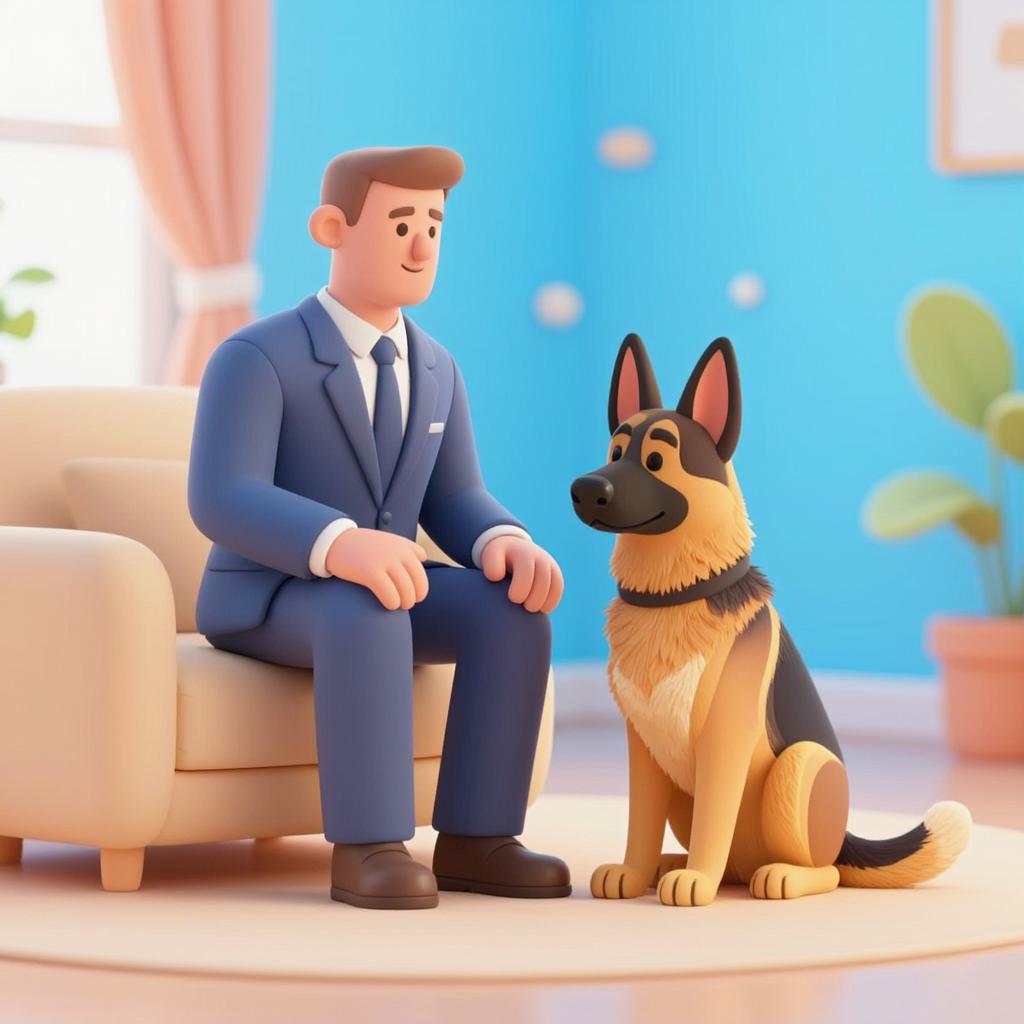}{i2L-Z-Image} &
        \imgcellb{0.23\textwidth}{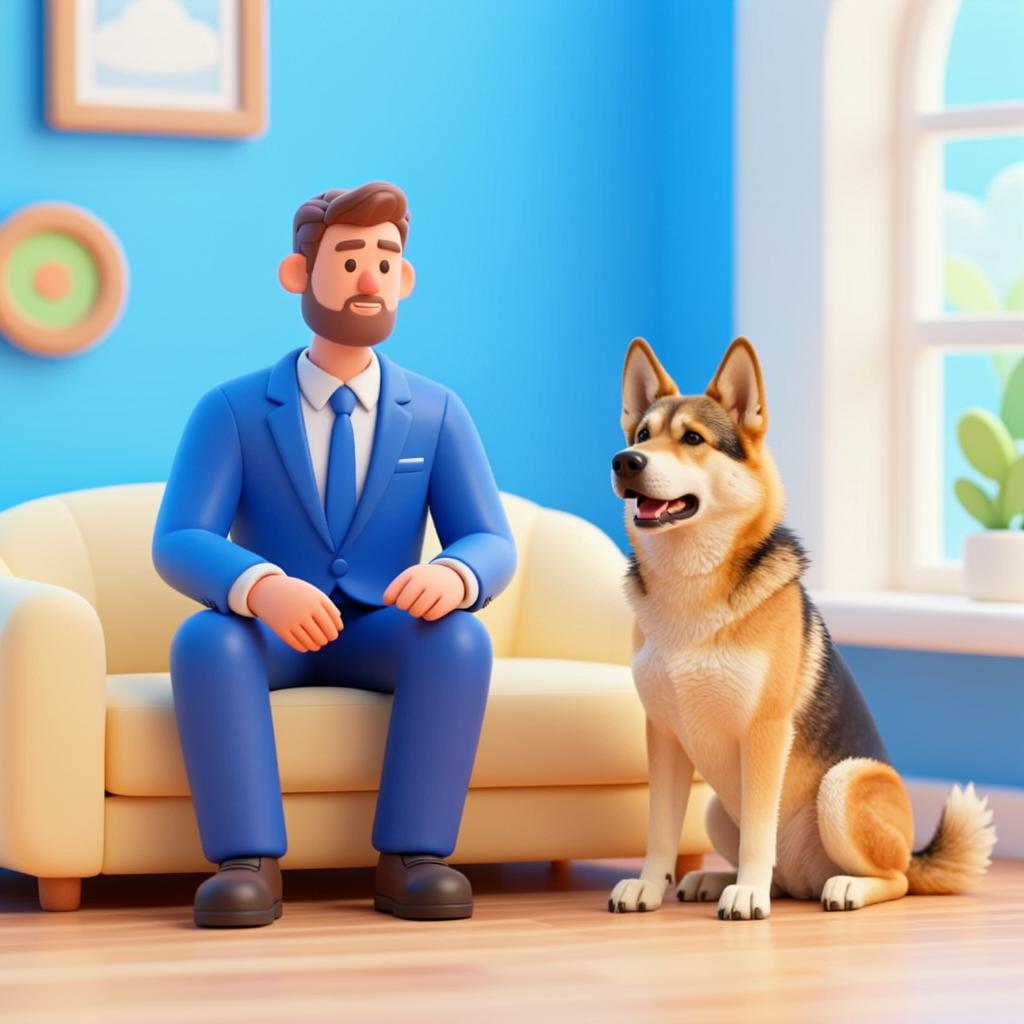}{i2L-FLUX.2} &
        \imgcellb{0.23\textwidth}{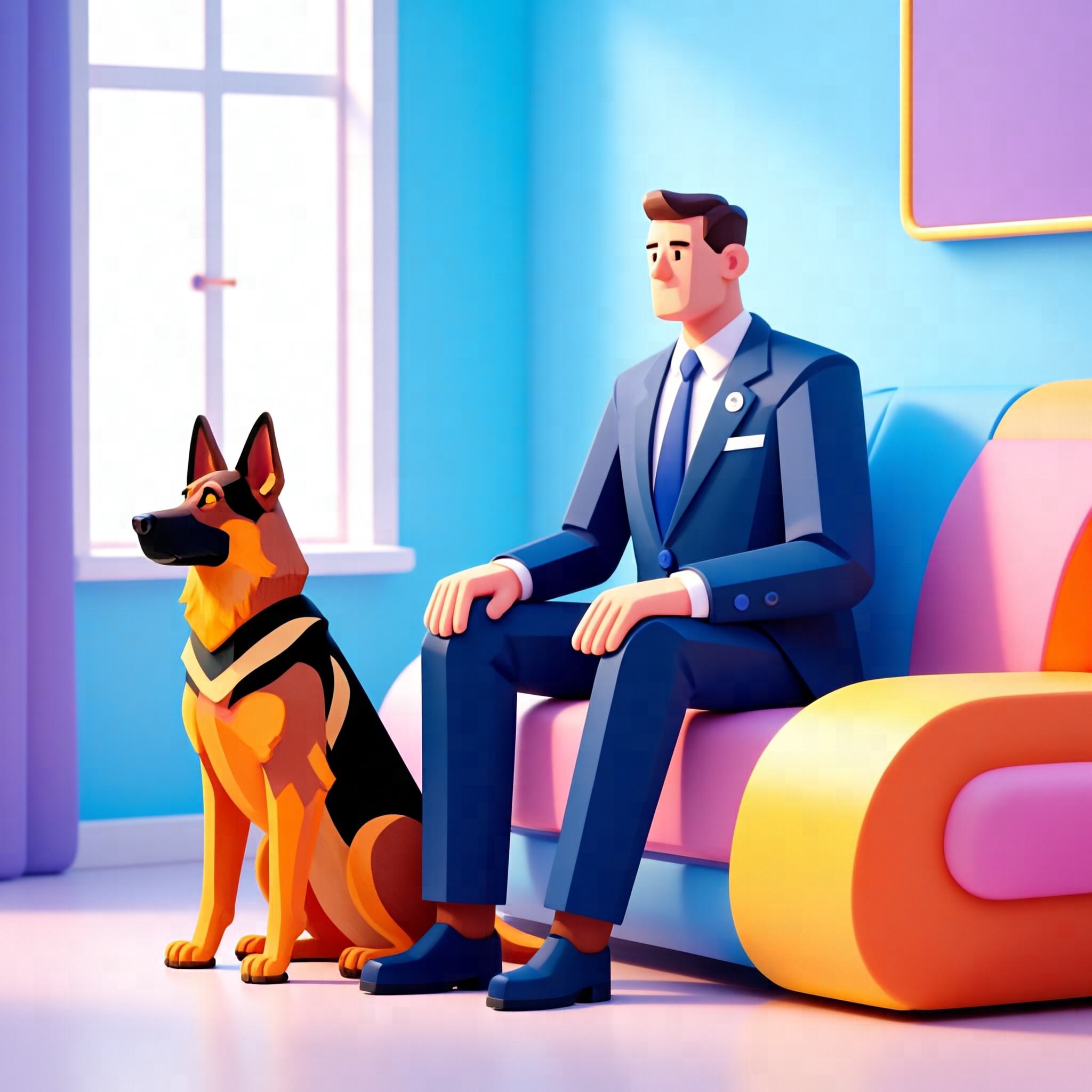}{i2L-Hidream-O1}
    \end{tabular}

    \vspace{0.15em}
    \begin{tabular}{cccccccc}
        \imgcell{0.112\textwidth}{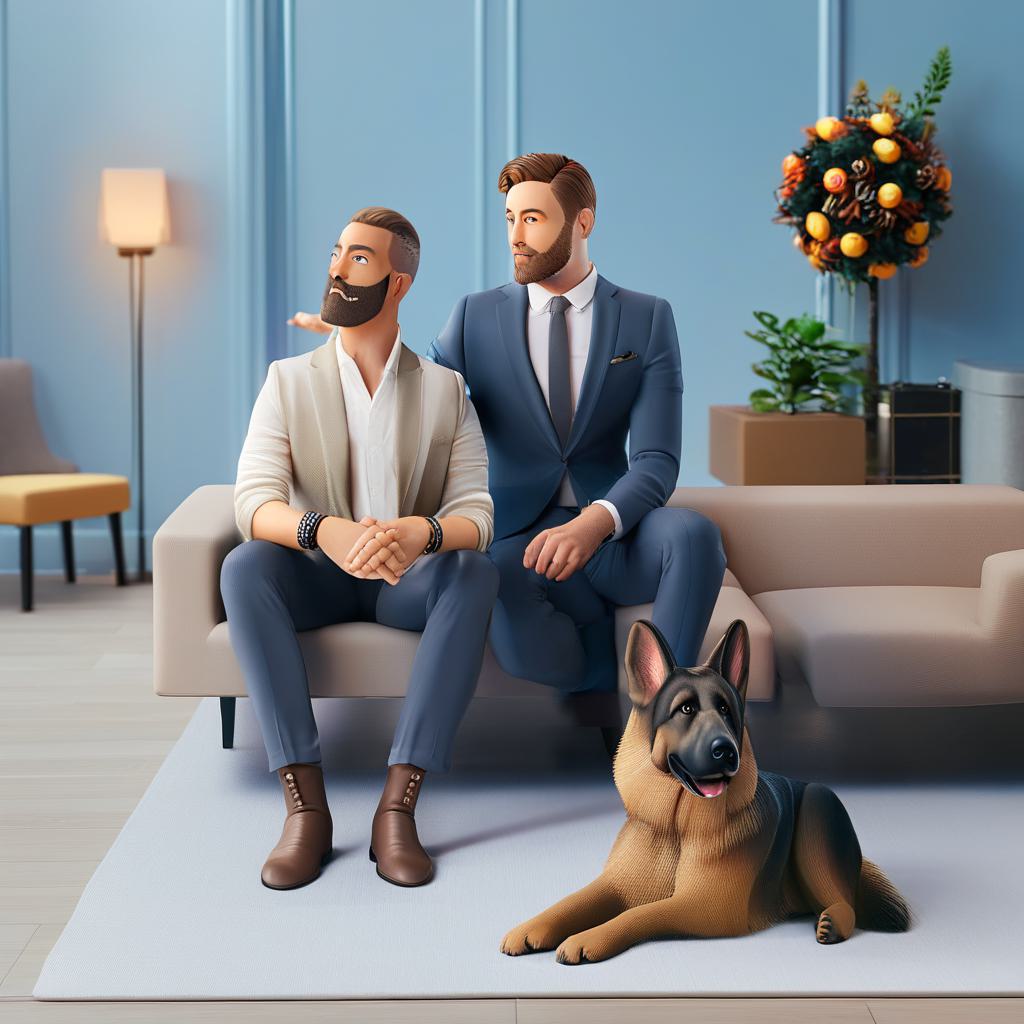}{StyleCrafter} &
        \imgcell{0.112\textwidth}{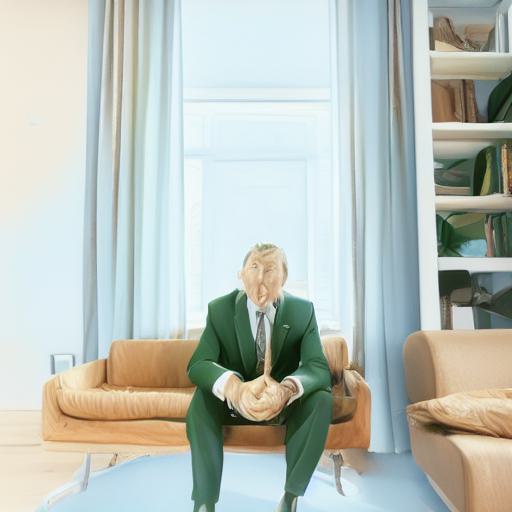}{StyleID} &
        \imgcell{0.112\textwidth}{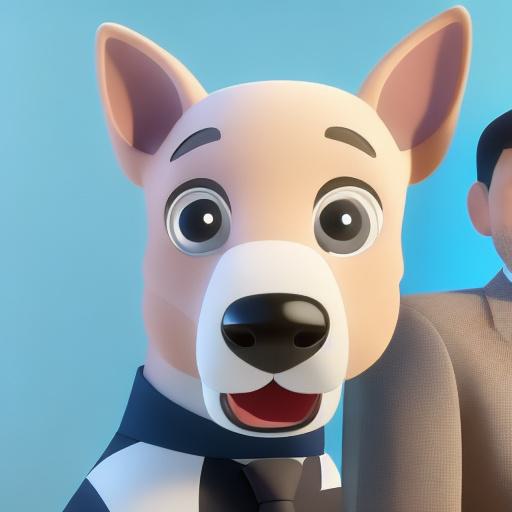}{ControlNet} &
        \imgcell{0.112\textwidth}{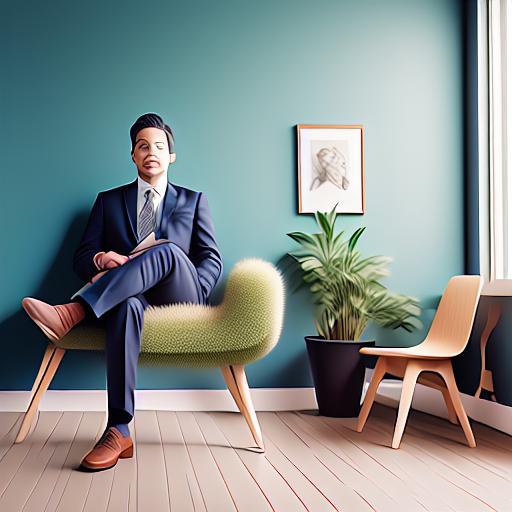}{DEADiff} &
        \imgcell{0.112\textwidth}{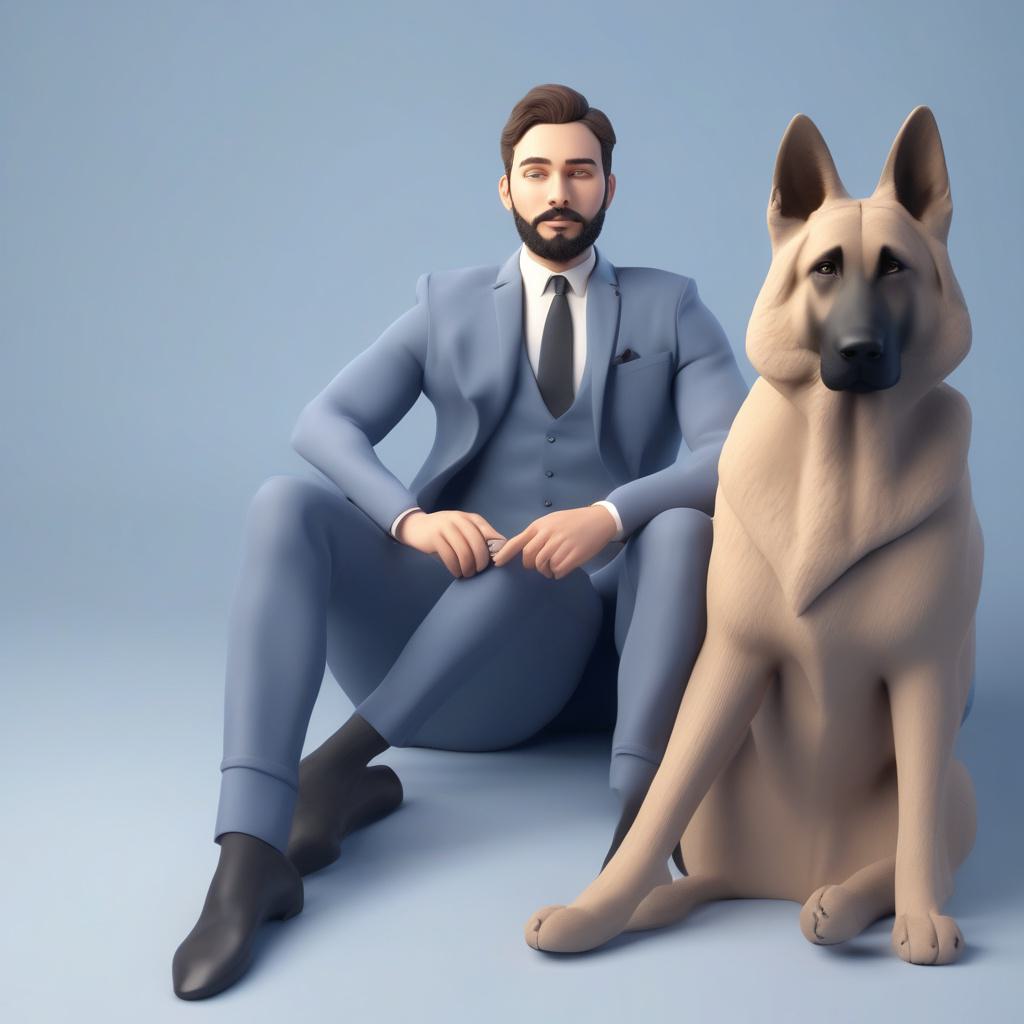}{InstantStyle} &
        \imgcell{0.112\textwidth}{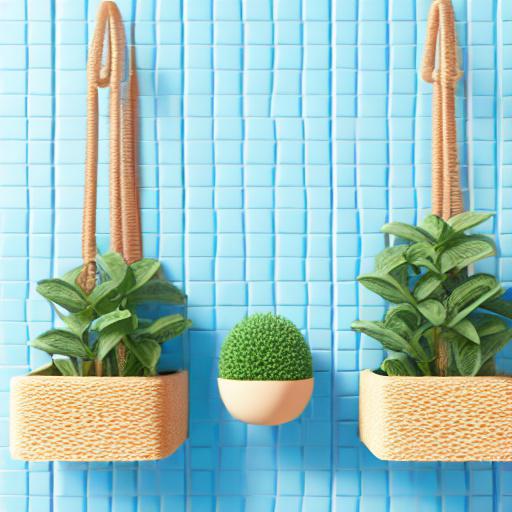}{IP-Adapter} &
        \imgcell{0.112\textwidth}{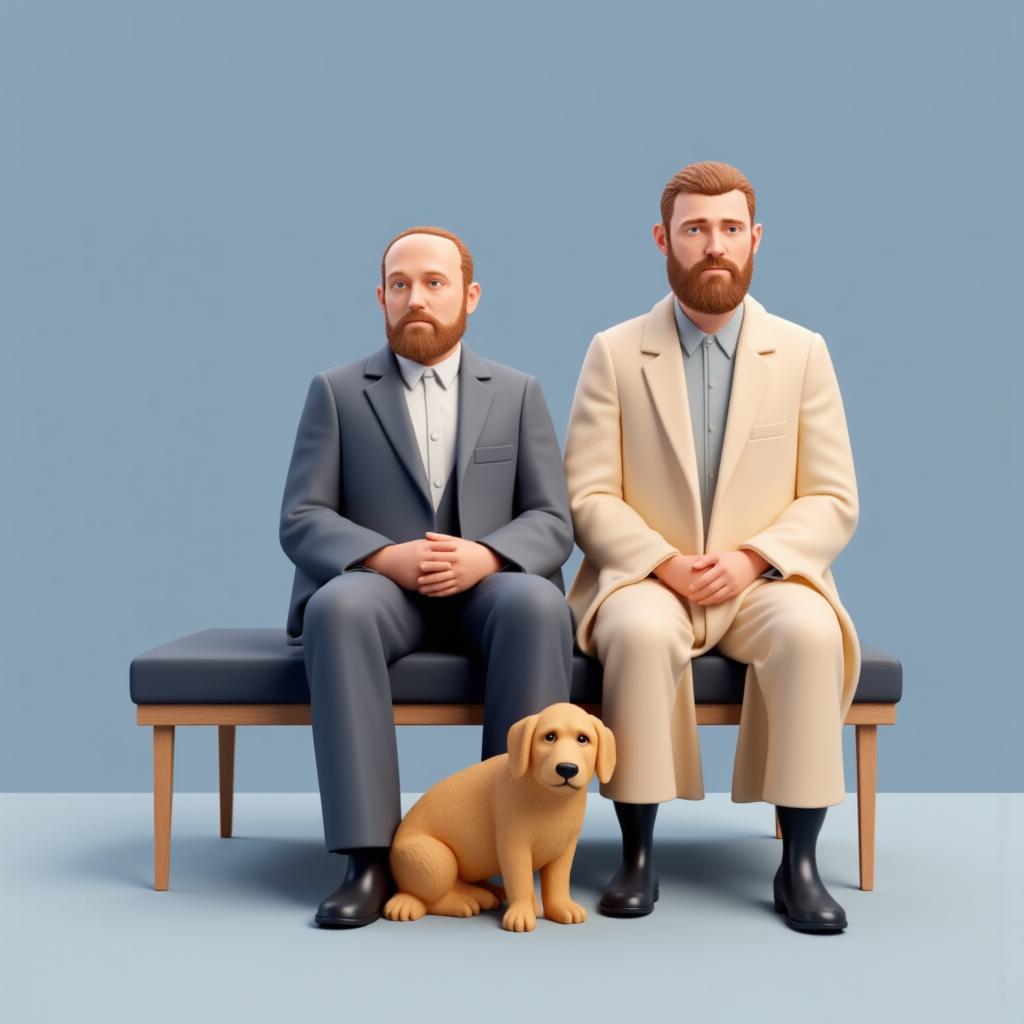}{IP-Adapter-FLUX} &
        \imgcell{0.112\textwidth}{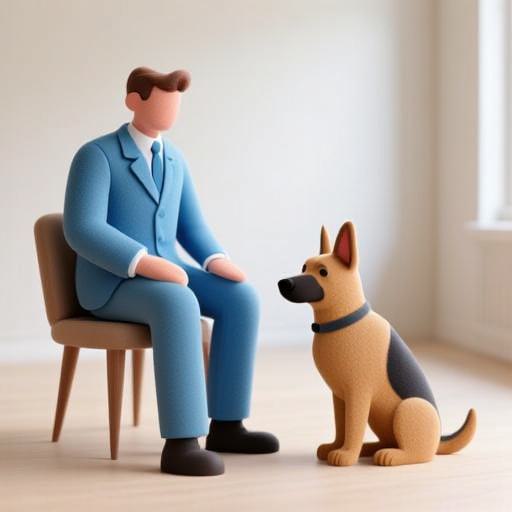}{MegaStyle-FLUX}
    \end{tabular}%
    \caption{Qualitative comparisons on three visualization groups. The prompts are ``Man in suit stands holding helmet near airplane,'' ``Bride and man walk through arched gateway,'' and ``Man in suit sits with German Shepherd indoors.''}
    \label{fig:visual_comparison}
\end{figure*}

\begin{table*}[htbp]
    \centering
    \small
    \setlength{\tabcolsep}{2.5pt}
    \begin{tabularx}{\textwidth}{l*{8}{>{\centering\arraybackslash}X}}
        \toprule
        Method & CLIP-Text & CLIP-Style & Aesthetic & PickScore & ImageReward & HPSv2 & HPSv3 & Overall \\
        \midrule
        StyleCrafter & 31.46 & 19.26 & 5.88 & 20.48 & 0.18 & 26.16 & 3.58 & -0.07 \\
        StyleID & 31.44 & 20.69 & 5.31 & 20.25 & -0.31 & 25.30 & 0.02 & -0.57 \\
        ControlNet & 22.08 & 21.32 & 5.83 & 19.29 & -1.40 & 23.93 & -1.56 & -0.82 \\
        DEADiff & 29.21 & 19.44 & 6.07 & 20.55 & -0.14 & 26.15 & 3.23 & -0.04 \\
        InstantStyle & 30.90 & 22.65 & 6.08 & 20.70 & 0.10 & 26.15 & 3.71 & 0.27 \\
        IP-Adapter & 6.70 & 23.83 & 6.17 & 17.77 & -2.13 & 21.45 & -2.49 & -1.18 \\
        IP-Adapter-FLUX & 10.73 & 23.16 & 6.19 & 18.40 & -1.87 & 23.47 & 0.24 & -0.78 \\
        MegaStyle-FLUX & 33.13 & 22.99 & 5.92 & 21.08 & 0.84 & 26.43 & 4.42 & 0.43 \\
        \midrule
        i2L-Z-Image & \underline{34.29} & \underline{25.53} & \underline{6.26} & \underline{21.53} & \underline{1.19} & \underline{27.51} & 5.68 & \underline{1.02} \\
        i2L-FLUX.2 & 33.58 & \textbf{25.57} & \textbf{6.36} & \textbf{21.57} & 1.18 & \textbf{27.60} & \textbf{6.03} & \textbf{1.08} \\
        i2L-Hidream-O1 & \textbf{34.71} & 21.13 & 6.11 & 21.47 & \textbf{1.27} & 27.47 & \underline{5.74} & 0.66 \\
        \bottomrule
    \end{tabularx}
    \caption{Quantitative comparison across multi-aspect metrics. The best score in each column is bolded, and the second-best score is underlined. Overall is computed as the average of normalized metric scores.}
    \label{tab:quantitative_results}
\end{table*}

\subsection{Experimental Settings}

\paragraph{Backbones and training.}
We train i2L on three foundation text-to-image backbones: Z-Image \cite{zimage2025} \footnote{\url{https://modelscope.cn/models/Tongyi-MAI/Z-Image}}, FLUX.2 \cite{flux-2-2025} \footnote{\url{https://modelscope.cn/models/black-forest-labs/FLUX.2-klein-base-4B}}, and Hidream-O1 \cite{hidream2025} \footnote{\url{https://modelscope.cn/models/HiDream-ai/HiDream-O1-Image}}, with 2.0B, 1.9B, and 2.3B parameters, respectively. The training framework is built on Diffusion Templates \cite{duan2026diffusion}. For each backbone, we train an independent image-to-LoRA predictor while freezing both the SigLIP2 image encoder and the base text-to-image model. Each model is trained for approximately seven days on 8 NVIDIA A100 GPUs, using a learning rate of $1\times10^{-5}$ and a global batch size of 8. At inference time, i2L predicts a LoRA directly from the reference images and inserts it into the corresponding backbone without per-style optimization.

\paragraph{Dataset.}
We use MegaStyle-1M \cite{gao2026megastyle} as the training set. It contains approximately one million training tuples designed to provide style-consistent but content-diverse examples. This property is important because the model must learn visual style rather than copy objects or identities from the reference images. We use a super-resolution model \footnote{\url{https://modelscope.cn/models/PAI/Z-Image-Turbo-Fun-Controlnet-Union-2.1}} to improve the resolution of each image to $1024\times 1024$. We hold out 1,000 examples for validation.

\paragraph{Baselines.}
We compare against a broad set of style-transfer and reference-conditioning baselines, including StyleCrafter \cite{liu2023stylecrafter}, StyleID \cite{chung2024styleid}, ControlNet \cite{zhang2023controlnet}, DEADiff \cite{qi2024deadiff}, IP-Adapter \cite{ye2023ipadapter}, IP-Adapter-FLUX \cite{flux-ipa}, InstantStyle \cite{wang2024instantstyle}, and MegaStyle-FLUX \cite{gao2026megastyle}. These baselines cover adapter-based image prompting, training-free diffusion style injection, controllable generation, and diffusion models trained on style-oriented data. For each method, we use the official inference configuration when available, and otherwise use the closest compatible setting. For ControlNet, we use the ``shuffle'' \footnote{\url{https://modelscope.cn/models/lllyasviel/control_v11e_sd15_shuffle}} model. The validation set contains 1,000 held-out samples, each with one prompt and four input images. For methods that accept only one input image, such as StyleID, we randomly sample one of the four images.

\subsection{Visualization}

Figure~\ref{fig:visual_comparison} presents three qualitative comparison groups. In these examples, i2L preserves the reference style while generating sharp images that remain aligned with the prompt. StyleCrafter, StyleID, ControlNet, and DEADiff show weaker instruction following or lower visual quality. IP-Adapter and IP-Adapter-FLUX exhibit semantic contamination by transferring non-style information from the references to the output. InstantStyle improves over IP-Adapter and produces cleaner images, but its style alignment remains weaker than that of the i2L variants. MegaStyle-FLUX follows both the prompt and reference style reasonably well, although fine details are sometimes suppressed.

\subsection{Quantitative Results}

Style transfer does not have a single ground-truth target, making reconstruction metrics unsuitable. We therefore use a multi-aspect evaluation protocol. CLIP-Text measures alignment with the target prompt and evaluates content consistency using CLIP \cite{radford2021clip}. CLIP-Style uses the same image-text alignment model to measure consistency between generated images and style descriptions from MegaStyle-1M. Aesthetic estimates visual appeal using an aesthetic predictor trained on human preference data \cite{schuhmann2022laion}. PickScore \cite{kirstain2023pickscore}, ImageReward \cite{xu2024imagereward}, HPSv2 \cite{wu2023hpsv2}, and HPSv3 \cite{ma2025hpsv3} provide complementary human-preference signals. The Overall score is computed as the mean of normalized scores across all metrics.

Table~\ref{tab:quantitative_results} reports the quantitative results, with the best score in bold and the second-best underlined. i2L-FLUX.2 achieves the best Overall score, while i2L-Z-Image and i2L-Hidream-O1 remain competitive on complementary metrics. The i2L variants consistently outperform feature-injection baselines, supporting the benefit of representing style in LoRA weights rather than solely through conditioning features. The high CLIP-Text score of i2L-Hidream-O1 further indicates that predicted LoRAs preserve content controllability.

\subsection{Ablation Study}

\begingroup

\begin{figure*}[htbp]
    \centering
    \includegraphics[width=\textwidth]{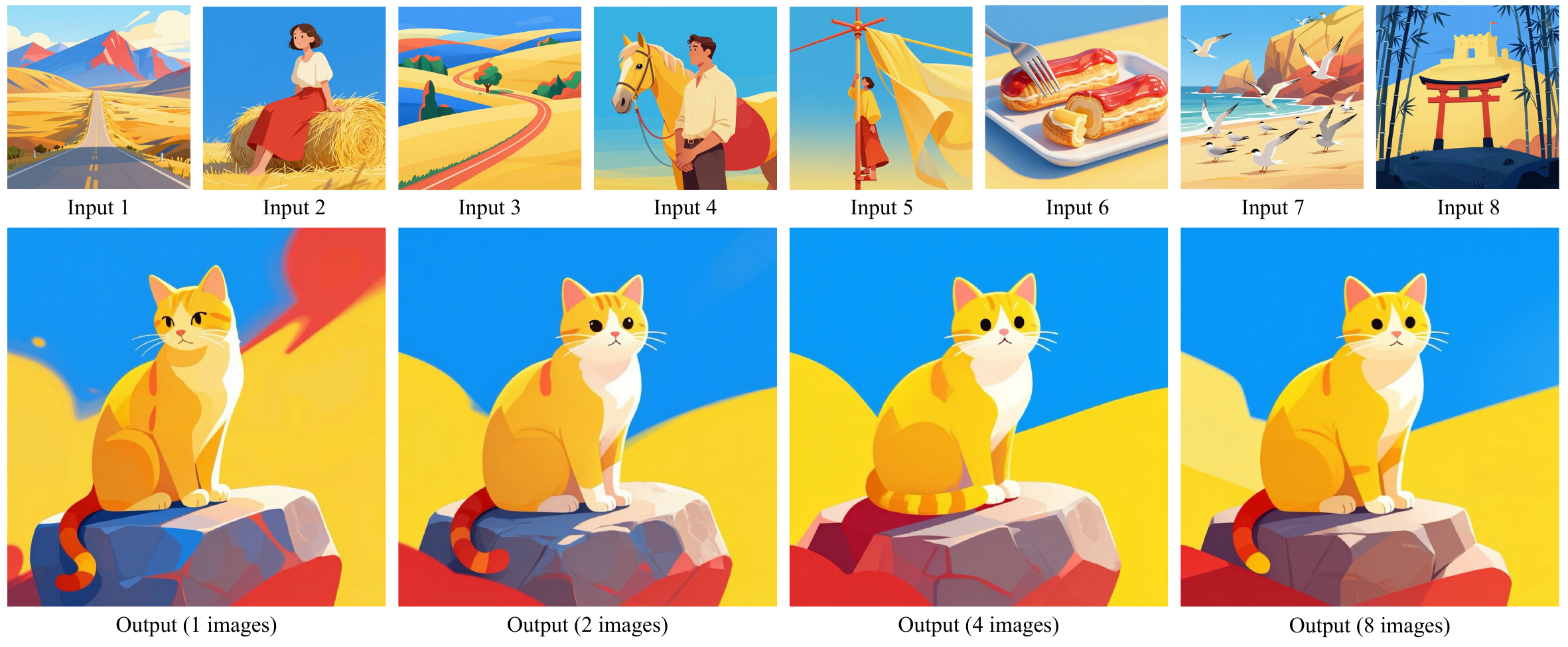}
    \caption{Ablation on the number of reference images. The first row shows eight style references, and the second row shows outputs generated with different reference-set sizes. Increasing the number of references provides a more reliable estimate of content-independent style and yields more stable LoRA predictions. The prompt is ``A cat is sitting on a stone'' and is shared by the following ablation figure.}
    \label{fig:ablation_num_images}
\end{figure*}

\begin{figure*}[htbp]
    \centering
    \includegraphics[width=\textwidth]{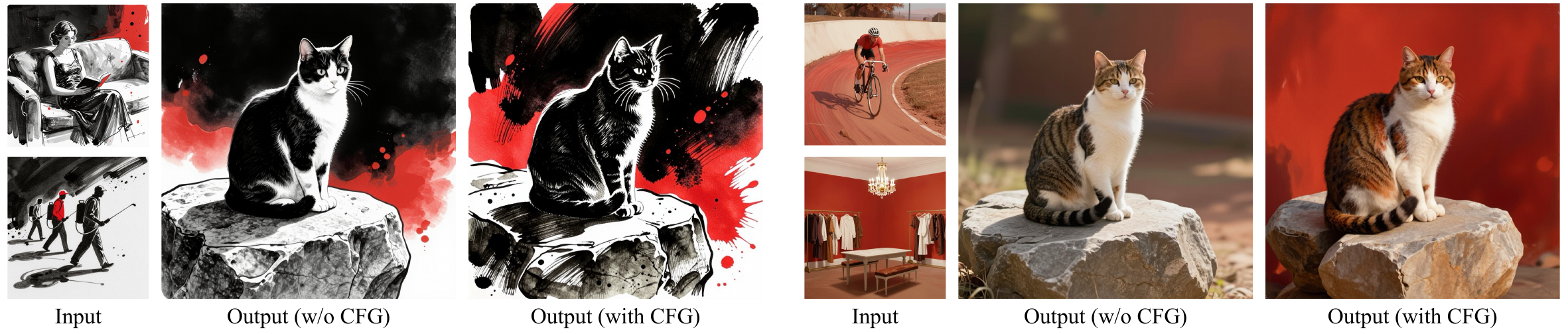}
    \caption{Ablation of asymmetric LoRA guidance. Input references are shown at the left side, followed by outputs without and with asymmetric LoRA guidance. Using different LoRAs in the positive and negative branches strengthens style adherence without additional training or per-style optimization.}
    \label{fig:ablation_cfg}
\end{figure*}

\begin{figure*}[htbp]
    \centering
    \includegraphics[width=\textwidth]{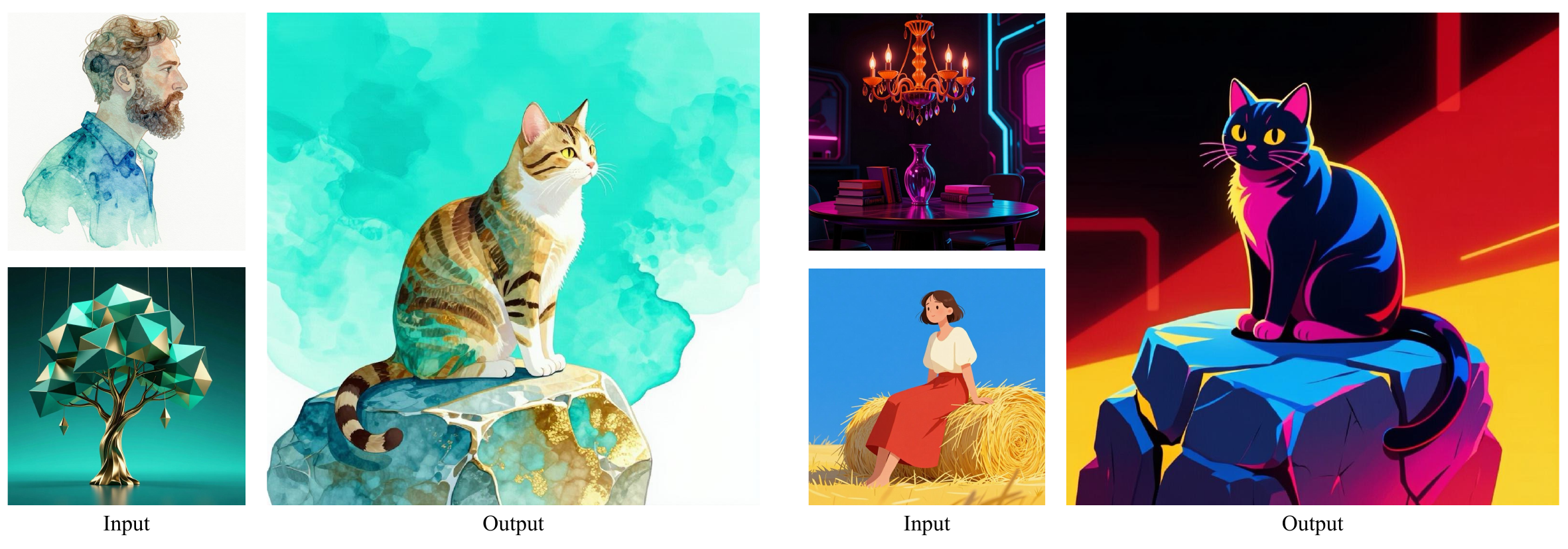}
    \caption{Multi-style fusion. The i2L model predicts one LoRA from multiple style references, enabling the generated image to inherit visual attributes from more than one style source.}
    \label{fig:fuse_multi_style}
\end{figure*}

\endgroup

\begingroup

\begin{figure*}[htbp]
    \centering
    \includegraphics[width=0.95\textwidth]{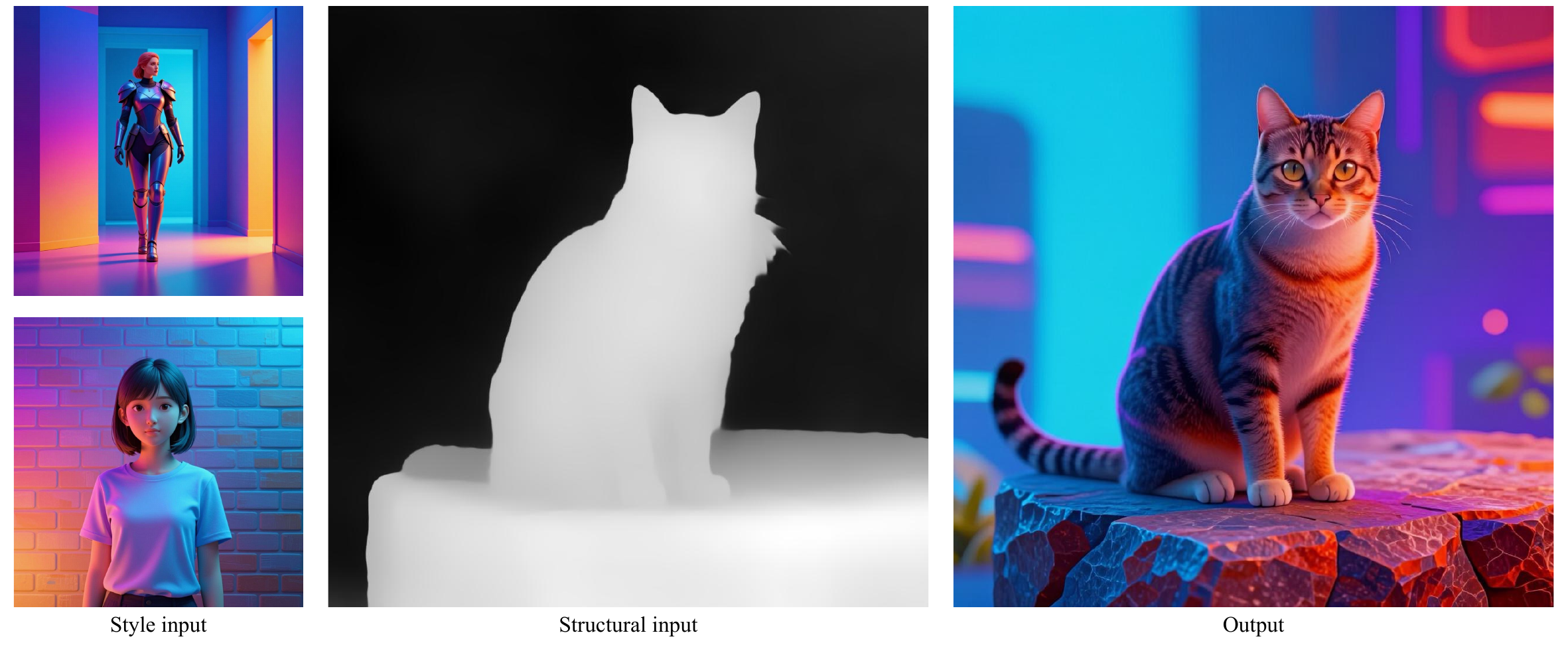}
    \caption{Composing i2L with ControlNet. The depth control determines spatial structure, while the predicted LoRA provides the reference style.}
    \label{fig:fuse_controlnet}
\end{figure*}

\begin{figure*}[htbp]
    \centering
    \includegraphics[width=0.95\textwidth]{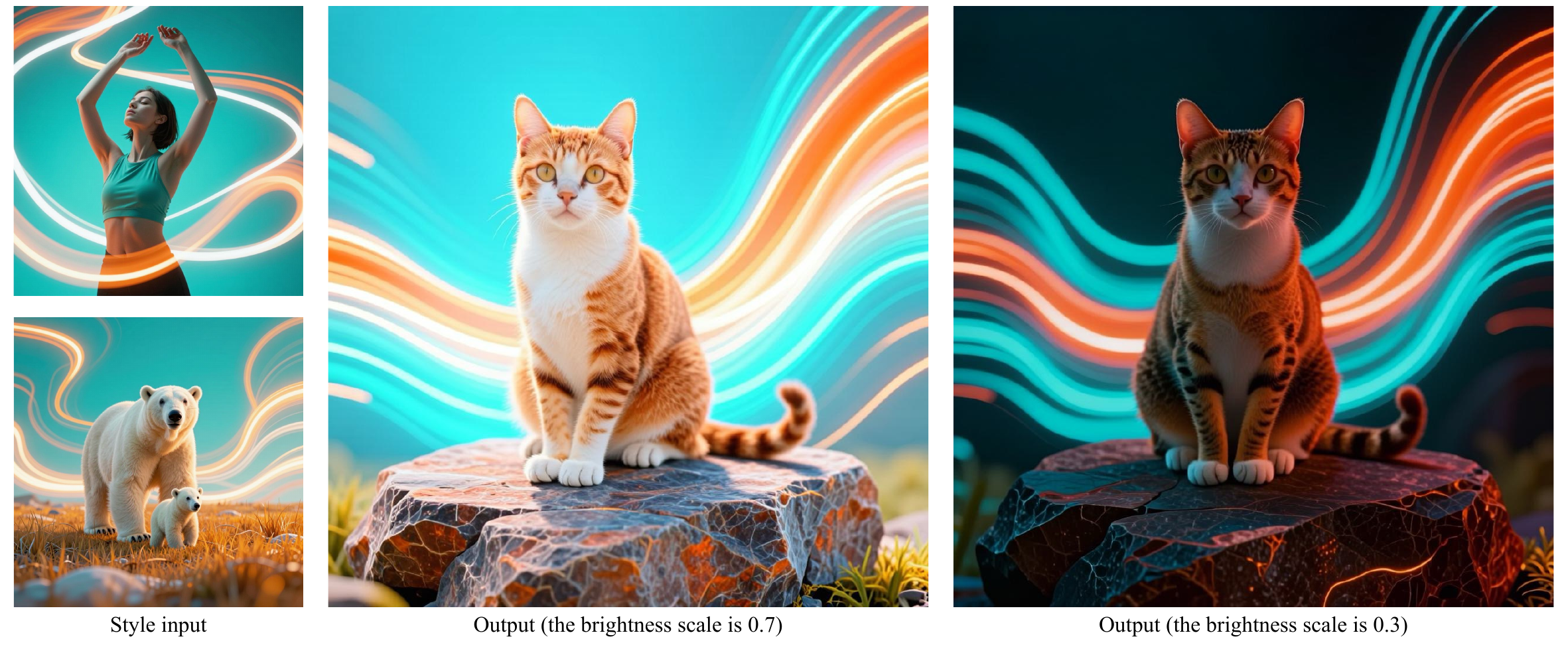}
    \caption{Composing i2L with AttriCtrl. The AttriCtrl module adjusts brightness while the predicted LoRA maintains the style extracted from the references.}
    \label{fig:fuse_attrictrl}
\end{figure*}

\begin{figure*}[htbp]
    \centering
    \includegraphics[width=0.95\textwidth]{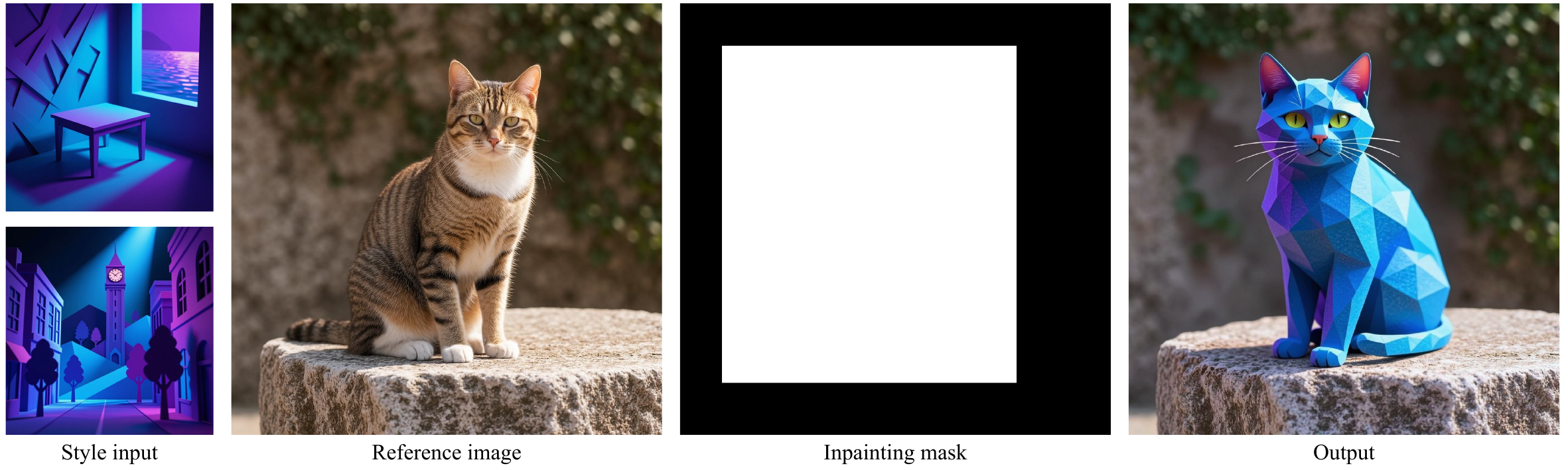}
    \caption{Composing i2L with inpainting. The edit follows the mask and reference image while inheriting the visual style encoded by the predicted LoRA.}
    \label{fig:fuse_inpaint}
\end{figure*}

\endgroup

\paragraph{Number of reference images.}
i2L supports a variable number of reference images because all references are encoded as image tokens and aggregated by the transformer before the LoRA queries decode the final weight update. Figure~\ref{fig:ablation_num_images} compares outputs generated with different reference-set sizes under the same prompt and sampling configuration. Across these settings, the generated images exhibit largely consistent visual styles, including similar color palettes, line quality, texture patterns, and rendering characteristics. This consistency indicates that i2L can extract and preserve the dominant style attributes even from very few reference images. Adding more references provides additional evidence for the same style and can further stabilize fine-grained details, but it does not fundamentally change the recovered visual language. This behavior suggests that the predicted LoRA captures style-level factors rather than relying on direct copying of individual reference images.

\paragraph{Asymmetric LoRA Guidance.}
Figure~\ref{fig:ablation_cfg} evaluates asymmetric LoRA guidance. Without this guidance strategy, the model can capture the dominant style but may under-emphasize subtle attributes such as fine outlines, low-contrast textures, and characteristic lighting. Asymmetric LoRA guidance applies the reference-image LoRA to the positive branch and a neutral gray-image LoRA, predicted by the same i2L network, to the negative branch. Because the two branches share comparable parameterization, their difference primarily reflects the style update induced by the reference images. The resulting guidance direction therefore amplifies stylization-specific effects rather than generic denoising behavior. Qualitatively, this strategy makes characteristic palettes, contours, and surface patterns more visible.

\subsection{Model Fusion Capability}

i2L outputs an explicit LoRA rather than transient conditioning tokens. The predicted weights can be stored, interpolated, reused across prompts, and combined with other modules through the standard LoRA interface of the base generator. This property makes the style representation modular: i2L controls appearance, while other inputs or control modules specify structure, illumination, masks, or editing constraints.

\paragraph{Fusing styles from multiple images.}
Because i2L accepts a set of reference images, it can fuse style cues from multiple inputs into a single LoRA in one forward pass. Figure~\ref{fig:fuse_multi_style} shows examples where two style images are provided jointly. Rather than selecting one reference, the generated image can combine compatible cues from both inputs, such as the palette of one image and the texture, line structure, or rendering pattern of the other. When the references share a coherent visual language, the fused LoRA produces a unified style instead of a spatial collage of source attributes. This behavior is consistent with transformer aggregation and LoRA decoding operating in a style-oriented representation space, where multiple references jointly determine the final weight update.

\paragraph{Composing with controllable generation.}
FLUX.2 supports a broad set of controllable generation modules. Using Diffusion Templates \cite{duan2026diffusion}, we combine i2L with ControlNet \cite{zhang2023controlnet} \footnote{\url{https://modelscope.cn/models/DiffSynth-Studio/Template-KleinBase4B-ControlNet}}, AttriCtrl \cite{chen2025attrictrl} \footnote{\url{https://modelscope.cn/models/DiffSynth-Studio/Template-KleinBase4B-Brightness}}, and inpainting \footnote{\url{https://modelscope.cn/models/DiffSynth-Studio/Template-KleinBase4B-Inpaint}}. In these pipelines, i2L supplies the style LoRA, while the external module controls structure, brightness, or editable regions. For ControlNet, the spatial condition determines the global geometry and pose, and the predicted LoRA transfers the reference appearance onto the controlled structure. For AttriCtrl, the brightness adjustment changes the image attribute while the LoRA maintains the reference-specific palette and texture. For inpainting, the mask and context define the edited region, and the predicted LoRA helps the inserted content remain stylistically consistent with the references. Figures~\ref{fig:fuse_controlnet}--\ref{fig:fuse_inpaint} show that predicted LoRAs remain effective under additional spatial or semantic constraints, making them practical style modules for complex generation workflows.

\section{Conclusion}

We have presented i2L, an image-to-LoRA framework that amortizes style LoRA training into a single model forward pass. Instead of treating a reference image as an external condition, i2L predicts explicit LoRA weights that directly modulate a frozen text-to-image generator. This design provides adapter-like efficiency while retaining the style internalization and composability of LoRA-based personalization. The query-based transformer and compressed decoding heads align the prediction problem with the row-and-column structure of LoRA matrices, allowing the framework to scale across modern diffusion backbones.

Experiments on Z-Image, FLUX.2, and Hidream-O1 show improved style fidelity, prompt consistency, and perceptual quality. Compared with reference-conditioned baselines, i2L preserves the visual language of the input style more reliably while reducing semantic leakage from reference images. Ablation studies show that multiple reference images yield more stable style estimates and that asymmetric LoRA guidance strengthens style adherence without additional training. Because i2L outputs a standard LoRA, the predicted style can also be fused across multiple references and composed with controllable-generation modules such as ControlNet, AttriCtrl, and inpainting. Overall, predicting generator weight updates from images offers a practical route to fast, high-fidelity, and controllable style transfer.

\section*{Acknowledgments}

We thank the open-source community for supporting and contributing to the LoRA model ecosystem. Through i2L, we aim to explore a new paradigm for LoRA model generation. We also acknowledge GPT for writing assistance during the preparation of this manuscript.

\bibliography{references}

\end{document}